\documentclass[11pt]{article}

\usepackage[preprint]{acl}

\usepackage{times}
\usepackage{latexsym}

\usepackage[T1]{fontenc}

\usepackage[utf8]{inputenc}

\usepackage[letterspace=80]{microtype}
\usepackage{newtxmath}

\usepackage{inconsolata}

\usepackage{graphicx}

\input{preamble}

\title{\TASKNAME: Evaluating Complex Reasoning \linebreak[1] via the Recognition of Languages In-Context} 

\author{%
  Jackson Petty\LING\Thanks{Correspondence to \texttt{research@jacksonpetty.org}.} \quad Michael Y. Hu\CDS \quad Wentao Wang\CDS \\ \bfseries Shauli Ravfogel\CDS \quad William Merrill\CDS \quad Tal Linzen\LING\CDS \\[1ex]
  Department of Linguistics\LING \quad Center for Data Science\CDS\\
  New York University
}

\begin{document}

\maketitle 

\begin{figure*}[ht!]
    \centering
    \includegraphics[width=0.75\linewidth]{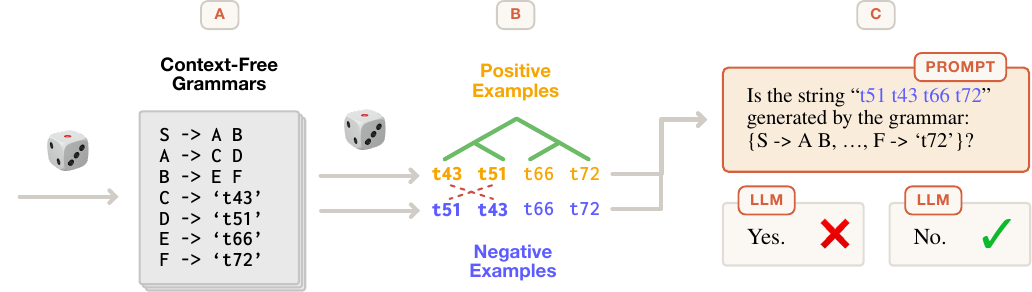}
    \caption{{\TASKNAME} is an evaluation framework for complex reasoning, where we \textbf{(a)}~stochastically generate context-free grammars of given complexities, \textbf{(b)}~sample positive and negative strings for each grammar, and \textbf{(c)}~prompt models to classify whether the strings are generated by those grammars.}
    \label{fig:concept}
\end{figure*}

\begin{abstract}
Large language models (LLMs) are increasingly used to solve complex tasks where they must retrieve and compose many pieces of in-context information in long reasoning chains. For many real-world tasks it is hard to accurately gauge how model performance and strategy change as task complexity grows. To evaluate models' complex reasoning capability in a scalable and verifiable way, we introduce {\TASKNAME} (Recognition of Languages In-Context), a framework that evaluates an LLM's ability to decide whether a given string belongs to the context-free language (CFL) generated by a grammar presented in-context. CFL recognition allows us to modulate the intrinsic complexity of the problem by varying grammar size and string length and translate this asymptotic complexity into predictions for ideal LLM performance. We find that even the most advanced reasoning models perform poorly on RELIC, not only failing to appropriately scale their inference compute to keep pace with task difficulty, but even reducing the number of reasoning tokens they use as task complexity increases. We find that these decreases in compute accompany changes in reasoning strategy, as models move from identifying and implementing algorithmic solutions to guessing. For models whose full completions go uninspected, this manifests as ``quiet quitting'' on hard tasks.

\bgroup \hfill
\textbf{Code:} \url{https://jpetty.org/relic} \hfill\null
\egroup
\end{abstract}

\section{Introduction}

Large language models (LLMs) are increasingly used to solve problems ``zero-shot,'' using only a task specification provided in the context window, without labeled examples or additional training. As these problems grow in complexity, models must reason through increasingly many interdependent steps and reference more and more pieces of in-context information; for instance, even a simple question like ``using my credit card statement, tell me if I spent more money in 2025 than I did on average in the past three years'' (inspired by \citealt{zhou-2024-webarena}) requires an LLM to retrieve many different items from its context and perform multiple operations in a particular sequence (in this example, averaging them and comparing the outputs). 

Formal domains, where solutions can be verified by an external system, are a valuable lens into LLMs' performance on complex reasoning tasks \citep{lambert-2025-tulu}. By studying reasoning problems in these domains, we can leverage our analytic understanding of the complexity of the formal problem to derive predictions for an optimal reasoner's performance on each instance of the task, and compare these predictions to the LLMs' empirical performance in a precise quantitative way.

Here, we propose the \textbf{RE}cognition of \textbf{L}anguages \textbf{I}n-\textbf{C}ontext (\textbf{RELIC}) framework, which applies this approach to \emph{language recognition}, a well-studied paradigm from theoretical computer science and computational linguistics. This paradigm consists of determining whether a given (formal) grammar generates a given string. In RELIC, an LLM is prompted with an arbitrary grammar and a string drawn from that grammar's terminal symbols and is asked to accept or reject the string based on the rules of the grammar. 

We focus on language recognition for two complementary reasons. First, this problem is well-studied in a classical algorithmic setting, allowing us to
predict how varying the hyperparameters which drive intrinsic task complexity, such as grammar size and string length, may affect performance.

Second, language recognition, though formally defined, has connections (\cref{sec:llm-reasoning}) to fuzzier `real-world' reasoning capabilities which are harder to study in their own right and which do not come equipped with precise controllability, such as \emph{instruction following} \citep{wei-2022-finetuned,finlayson-2022-what-a} and \emph{in-context language translation} \citep{tanzer-2023-benchmark-a}. Like these problems, RELIC requires sophisticated use of context: multi-hop retrieval of self-referential pieces (retrieving additional pieces based on ones already retrieved); composition of these pieces into candidate solutions; and a search over the candidate space for valid solutions. It thus serves as a formal analogue to these fuzzier tasks while bringing the benefit of easy verification.

In this work we instantiate the RELIC paradigm to explore for the first time how well LLMs can recognize \emph{context-free languages} (CFLs) in-context. CFLs are a class of formal languages designed to mimic the compositional structure of both programming and natural languages \citep{chomsky-1963-algebraic}, the two primary interaction media for LLMs. The asymptotic complexity of CFL recognition (\cref{sec:complexity-theory}) is within the capacity of transformers with chains of thought that can vary in length \citep{greenlaw-1991-compendium,merrill-2023-parallelism}; in particular, to solve the problem in all cases, the length of the chain of thought must be allowed to grow linearly in the size of the grammar and superlinearly in the length of the string. But this is only a theoretical possibility: it is an empirical question whether currently available LLMs, most of which are able to grow their chains of thought in this way, are able to solve this problem in practice.

We create a framework for stochastically generating novel context-free grammars (CFGs) and drawing positive and negative samples from these grammars. Using this framework, we construct a benchmark of grammars of varying size (up to 500 nonterminal productions) and accompanying sets of positive and negative example strings (of lengths of up to 50 symbols). We evaluate a range of LLMs on this benchmark, including smaller open-weight models (Gemma~3 and a distilled version of DeepSeek-R1) and proprietary models from OpenAI (\texttt{o3}, \texttt{o4-mini} and the \texttt{gpt-4.1} family). 

To preview our findings, most models perform well on small grammars and short strings, indicating a general understanding of the task; but all models, including the most advanced reasoning models, fall to near-chance accuracy on the larger grammars, which are still orders of magnitude smaller than the grammars needed to define commonly-used programming languages or approximate human languages. Qualitative analysis of the chain-of-thought (CoT) traces shows that in many cases models increasingly rely on incorrect heuristics; quantitatively, the lengths of CoTs do not grow as needed for the correct strategy to perform the task. For ``closed reasoning'' models that don't return full reasoning traces---like \texttt{o4-mini} and \texttt{o3}---this manifests as ``quiet quitting'' where models silently shift to prefer invalid reasoning strategies as problems get harder. 

These results indicate that LLM reasoning is brittle, even among state-of-the-art reasoning models. We conclude that there is substantial headroom for improvement on RELIC and, by extension, on the fuzzy problems for which language recognition is a formal analogue. The observation that models ``quiet quit'' motivates finding ways to ensure that LLMs stay faithful to valid reasoning strategies while mitigating the associated compute costs. More generally, these results illustrate the potential for using formal languages to produce evaluations that are resistant to data contamination, and whose difficulty can be precisely scaled as LLMs' capabilities improve.

\section{Background}

\subsection{LLMs and Complexity Analysis} \label{sec:complexity-theory}

We are inspired by a line of work that considers how intrinsic complexity measures for formal tasks determine how efficiently a model can solve the tasks based on the model's architecture, hyperparameters, and the number of test-time compute tokens. Prior work has examined transformers through the lens of circuit complexity, establishing that transformers without CoT (i.e., without the ability to produce intermediate tokens before responding) are in the complexity class $\TC^0$ \citep{merrill-2023-parallelism}, while those with linearly-many CoT steps are in the complexity class $\NC^1$ \citep{merrill-2024-expressive}. 

The recognition of membership in a context-free grammar without $\varepsilon$-productions is known to be in the complexity class $\AC^1$ and to be $\NC^1$-hard\footnote{$\TC^0$ problems are those decidable by circuits of constant depth and polynomial size, containing binary and majority gates; $\NC^1 \supseteq \TC^0$ problems are decidable by circuits of logarithmic depth and polynomial size, with each gate having at most two inputs; $\AC^1 \supseteq \NC^1$ problems are decidable by circuits of logarithmic depth and polynomial size, but with no fan-in restrictions on gates.} \citep{ruzzo-1980-tree-size,venkateswaran-1991-properties,greenlaw-1991-compendium}. Standard context-free parsing algorithms run in time $\Theta(G\cdot\ell^3)$ with little likelihood of significant improvement (where $G$ is the number of rules in the grammar and $\ell$ is the length of the string; \citealt{lee-2002-context-free}). Translating this runtime-complexity into steps of inference \citep{merrill-2024-expressive}, this means that transformers with CoT can, in theory, solve RELIC, but at a steep cost: the number of CoT steps required is bounded between at least $\Omega(G\cdot\ell^{1.7})$ and at most $O(G\cdot\ell^6)$. Since inference compute is, in practice, a limited quantity, we then predict that there are complexity bounds on both grammars and strings beyond which transformer LLMs will be unable to perform above chance.

\subsection{LLMs and Context Use} \label{sec:context-use}

In addition to the ability to execute algorithms in line with theoretical measures of their difficulty, to solve complex tasks provided in a prompt, LLMs require the more prosaic ability to effectively use information provided in-context.

Early work on LLMs' context-use ability focused on retrieval tasks like ``needle-in-the-haystack'' evaluations where models are tasked with retrieving specific pieces of information provided in-context \citep{kamradt-2023-gkamradt/llmtest_needleinahaystack,hsieh-2024-ruler,bohnet-2024-long-span,zhang-2024-inftyBench,arora-2024-zoology}. More difficult tests for context use introduce elements of \emph{referentiality} and \emph{compositionality}\footnote{We mean compositionality in a broad sense of combining atomic primitives in some well-defined manner; the more specific sense of deriving the \emph{meaning} of expressions from their constituent parts is implicated in question answering \citep{dziri-2023-faith,press-2023-measuring} and semantic parsing \citep{kim-2020-cogs}.} (piecing together multiple pieces of retrieved information). \citet{kim-2020-cogs} and \citet{press-2023-measuring}, \emph{inter alia}, demonstrate the challenge that compositional reasoning poses for language models in both toy and real-world settings. 

More recent work in this direction includes \citet{wu-2025-transformers}, which provides an account of how language models learn to implement variable-binding, a necessary component of compositional reasoning; NeedleBench \citep{li-2025-needlebench}; and Michelangelo \citep{vodrahalli-2024-michelangelo}. Like RELIC, Michelangelo consists of synthetically generated tasks, which make it possible to control the complexity of the task and address dataset leakage issues. Our work is motivated by similar concerns but studies the substantially more challenging task of language recognition---a setting which is harder both in the formal complexity-theory sense addressed above, and because unlike these datasets RELIC is characterized by \emph{non-monotonicity} (the order in which pieces of the context must be retrieved or modified may not be the same as their appearance order) and \emph{search} (multiple candidate solutions must be constructed from retrieved pieces and evaluated to explore the solution space; \citealt{yao-2023-tree-a, yang-2023-large}). 

\subsection{LLMs and Reasoning} \label{sec:llm-reasoning}

Poor performance of pure language models on tasks requiring reasoning motivated the development of increasingly sophisticated approaches to tackle complex problems, notably by scaling the amount of compute models expend at inference time (test-time compute), first through methods like chain-of-thought prompting \citep{wei-2022-chain-of-thought-a}, and later through iterated self-reflection (\citealt{openai-2024-learning}) in reasoning language models like OpenAI's o1, o3, o4-mini, and GPT-5 \citep{openai-2024-openai,openai-2025-openai,openai-2025-gpt-5}; Gemini Thinking \citep{google-2025-gemini}; Claude Thinking \citep{anthropic-2025-claudes}; and DeepSeek-R1 \citep{deepseek-ai-2025-deepseek-r1}. These approaches have led to broad improvements in performance across a wide array of static reasoning benchmarks. Yet recent work has also noted the brittleness in reasoning performance \citep{shojaee-2025-illusion-a}, where even the best models fail to implement consistent algorithmic solutions in many cases or to generalize to new instances or kinds of tasks.

Two classes of problem of particular relevance to RELIC are \emph{instruction following} and \emph{in-context translation}. Regarding the former, `base' LLMs are typically fine-tuned to follow in-context instructions, and can generalize this skill to new instruction sets not seen during training \citep{sanh-2022-multitask,wei-2022-finetuned}. 
Language recognition was first proposed as a formal analogue to instruction following by \citet{finlayson-2022-what-a}, who note that grammars function as instruction sets for the task of (regular) language recognition. RELIC expands on work in this direction by focusing on a richer class of languages, CFLs, which have a higher level of intrinsic complexity than regular languages do and consequently are better able to model the kinds of compositional structures present in both natural and programming languages \citep{chomsky-1959-certain,chomsky-1963-algebraic}. Unlike work on LLM language recognition where the LLMs were trained for the specific task or on specific grammars \citep{bhattamishra-2020-ability,deletang-2023-neural,butoi-2025-training}, we examine the ability of general-purpose LLMs to accept or reject strings.

The second class of problems that is closely related to RELIC, \emph{in-context translation}, was first proposed by \citet{tanzer-2023-benchmark-a}. In this task, an LLM is prompted with a complete description of a language (e.g., a reference grammar or textbook) and is expected to use this description to translate into a low-resource language that was not included in its pretraining corpus. This experimental design, while innovative, suffers from the difficulty of verifying the correctness of low-resource natural language translations and leaves open the question of how LLM performance is impacted by the language description's complexity. Our work here begins to address some of these concerns. As an instance of language \emph{recognition}, RELIC is closely related to language translation, and is the formal analogue to the task of determining whether or not a translated sentence is \emph{grammatical}, independent from whether or not it is the correct translation. Very recently \citet{gupta-2025-randomly} has begun to explore in-context translation performance through formal means, but only for regular languages too simple to be good proxies for natural, human languages. 

\section{\TASKNAME: Recognizing Languages In-Context}
\label{sec:relic}

{\TASKNAME} is a framework for generating synthetic benchmarks: it can be used to generate new evaluation instances of increasing complexity that are extremely unlikely to have been observed by the model in training. In this paper, in addition to implementing this process in a codebase, we also generate a static benchmark using this process, and evaluate contemporary language models on the benchmark.

Following \citet{clark-2017-testing,clark-2018-syntheticpcfg}, we stochastically generate grammars which are parameterized by four values: the number of terminal symbols $\Nterm'$, nonterminal symbols $\Nnonterm'$, lexical production rules $\Nlex'$, and nonlexical production rules $\Nnonlex'$. We start by defining $\Nterm'$ symbols $\Sigma = \{ \tok{t}_1, \tok{t}_2, \dotsc \}$; and $\Nnonterm'$ symbols $V = \{\tok{NT}_1, \tok{NT}_2, \dotsc \}$. We then sample $\Nlex'$ lexical production rules $\tok{NT}_a \to \texttt{`$\tok{t}_b$'}$ from the set of pairs $V \times \Sigma$, and we sample $\Nnonlex'$ nonlexical production rules $\tok{NT}_a \to \tok{NT}_b ~~ \tok{NT}_c$ from the set of all triples $(\{\texttt{S}\} \cup V) \times V \times V$, where \texttt{S} is a privileged start symbol. The grammar is then trimmed to remove any nonlexical rules which do not lead to a lexical rule, and any lexical rules which are inaccessible from nonlexical productions. The result is a reduced context-free grammar $\mathbf{G}$ with at most as many terminals $\Nterm$, nonterminals $\Nnonterm$, lexical productions $\Nlex$, and nonlexical productions $\Nnonlex$ as the generating parameters; for analysis, we measure the grammar's complexity according to these values after reduction, and specifically define its size as $G = \Nlex + \Nnonlex$.

Using this framework, we generate a first static benchmark, which we refer to as \textbf{\TASKNAME-500}. We sample 200 grammars where all parameters ($\Nterm$, $\Nnonterm$, $\Nlex$ and $\Nnonlex$) are less than 500. To reduce the correlation between the four parameters, we oversample many grammars and filter to obtain the 200 with minimally-correlated parameters; we note that some parameter pairs, such as $\Nlex$ and $\Nterm$, are inherently correlated and it is not possible to completely decorrelate them (see~\cref{fig:grammar-correlations} for the correlation coefficients).

For each grammar, we aim to sample 10 positive strings $\mathbf{s}$ of lengths $1 \leq \ell \leq 50$ by treating each grammar as a probabilistic context-free grammar with uniform probability assigned to production rules sharing a left-hand symbol. We also generate 10 negative strings (i.e., strings which are not generated by the grammar) of each length: we sample negative strings from a unigram model over the terminal symbols $\Sigma^+$ and reject any strings which are parseable by the grammar.\footnote{Our codebase also includes the ability to sample negative strings adversarially by making minimal edits to positive samples. Since model performance on RELIC-500 is quite poor across the board, in our experiments we only report our results for the easier class of negative examples sampled from the unigram distribution, but for the purpose of evaluating more advanced models in the future, we encourage the use of the adversarial negative examples.} Not all randomly-generated grammars will be able to produce arbitrarily many strings of a given sequence length. We refer to the size of the set of generated examples relative to a defined goal of 1000 examples as the grammar's \emph{coverage}, and report it as a summary statistic of each grammar. For the experiments conducted here, the majority of grammars have over 90\% coverage; see \cref{sec:grammar-stats} for more details. Note that due to the sampling procedure, the distribution of lengths between positive and negative samples is not matched; see \cref{fig:sample-stats} for the ratios of positive to negative examples for each length $1 \leq \ell \leq 50$. In our analyses (\cref{sec:results-properties}), we rebalance the classes.

\paragraph{Data novelty \& contamination.} RELIC is designed to mitigate issues of data contamination by synthetically generating novel grammars rather than relying on a fixed dataset. Since the number of possible grammar-string pairs grows very rapidly in both the size of the grammar and the length of the string (see~\cref{sec:grammar-growth} for more details), the risk of ever sampling a new grammar which could have been seen by a model during training is vanishingly small.

\paragraph{Intended use.} {\TASKNAME} is designed as a \emph{zero-shot} evaluation of a model's ability to reason about complex in-context tasks \emph{without also using positive and negative exemplars}. Evaluating models in a few-shot setting could lead to higher accuracy, but any increases in accuracy could be due to heuristics that distinguish the two classes with some accuracy but fail to strictly test the model's ability to follow in-context instructions---for example, ones that rely on differences in the $n$-gram distributions in positive versus negative examples. 

\paragraph{Dataset \& code release.} For reproducibility and to facilitate the creation of new RELIC benchmarks, we release \TASKNAME-500 and the codebase to generate new grammars and examples at \url{https://jpetty.org/relic}.

\section{Experimental Methodology} \label{sec:exp-setup}

We evaluate eight LLMs on {\TASKNAME}-500. First, three models from OpenAI's GPT line that were the newest at the time of writing: \texttt{gpt-4.1-nano}, \texttt{gpt-4.1-mini}, and \texttt{gpt-4.1} (\citealt{openai-2025-introducing}). Second, we evaluate two reasoning models from OpenAI, \texttt{o4-mini} and \texttt{o3} \citep{openai-2025-openai} which are trained to generate a variable number of hidden tokens between the prompt and the final output. All evaluations of OpenAI's API-based models were carried out between 14 April 2025 and 1 May 2025. Third, we evaluate two models from Google's Gemma~3 family of open-weight instruction-tuned language models (\texttt{gemma-3-1b-it} and \texttt{gemma-3-4b-it}; \citealt{gemmaTeam-2025-gemma}). Finally, we evaluate an open-weight reasoning model, \texttt{DeepSeek-R1-Distill-Qwen-7B} \citep{deepseek-ai-2025-deepseek-r1}. See~\cref{sec:hyps} for more details on API costs, compute, and evaluation hyperparameters. 

For each grammar $\mathbf{G}$ and each string $\mathbf{s}$, we prompt the model with a description of the language recognition task, the grammar, and the string, and ask the model to classify the string according to whether or not it is generated by the grammar, as shown below.\footnote{Prompts and LLM responses are shortened for brevity and clarity by eliding ancillary text with [...].}

\begin{prompt}
You will be presented with a context-free grammar in Chomsky normal form 
and a string which may or may not be in the language defined by the given 
grammar. Your job is to determine whether or not the grammar generates the
provided string. You can use any reasoning strategy you like, but you must 
end your response with either `Yes' (if the string is generated by the 
grammar) or `No' (if it isn't.) 

Grammar:
\begin{verbatim}
```
S -> NT97 NT1
NT180 -> NT47 NT121
NT120 -> NT73 NT121
NT114 -> NT197 NT79
NT191 -> NT76 NT49
NT8 -> NT90 NT28
NT192 -> NT140 NT152
\end{verbatim}
\llmfiller 
\begin{verbatim}
NT171 -> 't59'
NT31 -> 't139'
NT172 -> 't28'
NT100 -> 't16'
NT187 -> 't158'
NT100 -> 't44'
```
\end{verbatim}

Here is the string you need to evaluate:

String: \verb|`t64`|.

Remember, end your response with either `Yes' or `No'.
\end{prompt}

\vspace*{0em}

We use a regular expression to determine whether the model classified the example as positive or negative; we classify the response as ``unknown'' if the model fails to offer a prediction. We evaluate \texttt{gpt-4.1-nano}, \texttt{gpt-4.1-mini}, \texttt{gpt-4.1}, and \texttt{o4-mini} on the full {\TASKNAME}-500 dataset; for \texttt{o3}, \texttt{gemma-3-1b-it}, \texttt{gemma-3-4b-it}, and \texttt{DeepSeek-R1-Distill-Qwen-7B}, we subsample each grammar's data to have at most two examples per length per type (instead of 10) due to the increased cost of running evaluations on these models.

\section{Experimental Results}

\subsection{Overall Performance}

\begin{figure*}
    \centering
    \includegraphics[width=\linewidth]{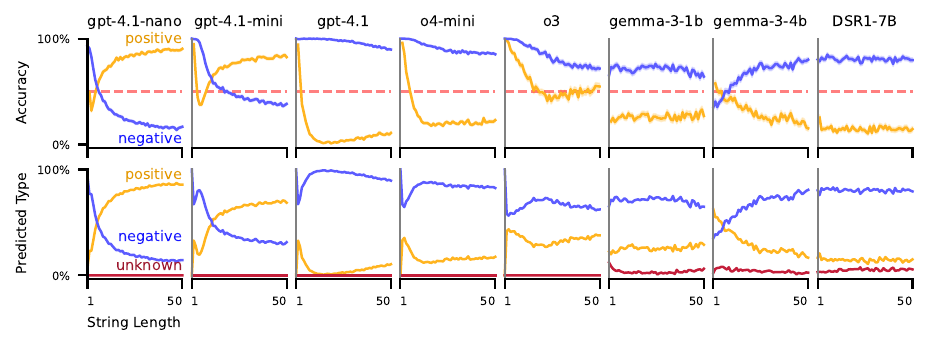}
    \caption{Accuracy on positive examples (that is, examples that can be generated by the grammar in question) and negative examples differs between models and across string lengths (\textbf{top row}); this is partially attributable to models' biases towards predicting one class (positive or negative) over another (\textbf{bottom row}).}
    \label{fig:acc-by-type-length}
\end{figure*}

In \Cref{tab:model_accuracy}, we report models' mean class-balanced accuracy and their macro F1 score (mean of the per-class F1 score); since models do not have access to the distribution from which positive and negative examples are drawn, class-balancing penalizes models for incorrect biases towards predicting one class (positive or negative) over another. Indeed, models' accuracy often differs significantly between positive and negative examples, indicating such a bias: \texttt{gpt-4.1-nano} and \texttt{gpt-4.1-mini} have a much higher accuracy on positive examples than they do on negative ones, while the reverse is true for \texttt{gpt-4.1}, \texttt{o4-mini}, and \texttt{o3}~(\cref{fig:acc-by-type-length}, top). 

\begin{figure}[!ht]
    \centering \small
    \begin{tabularx}{\linewidth}{
        X
        S[table-format=2.1(1.1)]
        S[table-format=2.1(1.1)]
    }
      \toprule
      \textbf{Model}                        & \textbf{Accuracy (\%)}       & \textbf{Macro F1}         \\
      \midrule
      \texttt{gpt-4.1-nano}        & 54.1(3.0)             & 44.4(0.7)     \\
      \texttt{gpt-4.1-mini}        & 63.4(1.9)             & 57.5(1.3)     \\
      \texttt{gpt-4.1}             & 53.9(4.5)             & 48.5(0.9)     \\
      \texttt{o4-mini}             & 59.2(3.5)             & 58.1(1.0)     \\
      \texttt{o3}                  & 70.4(1.9)             & 70.1(1.1)     \\
      \texttt{gemma-3-1b}          & 48.8(2.3)             & 30.9(0.3)     \\
      \texttt{gemma-3-4b}          & 48.7(2.1)             & 34.3(0.7)     \\
      \texttt{DSR1-7B}             & 47.9(3.3)             & 29.6(0.3)     \\
      \bottomrule
    \end{tabularx}
    \caption{Overall performance (class-balanced accuracy and macro F1 score) on {\TASKNAME}-500. Chance performance is 50\%. Performance is calculated first by grouping observations by grammar, example type, and example length, and then taking the mean. Errors reflect the standard error of the mean.}
    \label{tab:model_accuracy}
\end{figure} 

This difference in accuracy is mostly attributable to the individual models' bias towards one of the two classes (\cref{fig:acc-by-type-length}, bottom); while most models exhibit a strong tendency to classify short examples as negative, \texttt{gpt-4.1-nano} and \texttt{gpt-4.1-mini} increasingly classify examples as positive as they get longer, to the point that nearly all their predictions for examples of length $\ell=50$ are positive. \texttt{gpt-4.1}, by contrast, exhibits nearly the opposite tendency, classifying most examples as negative across all lengths.

\begin{figure*}
    \centering
    \includegraphics[width=\linewidth]{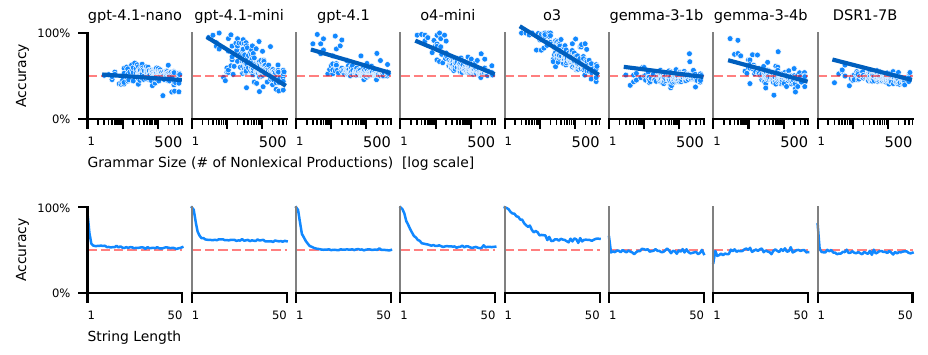}
    \caption{Models' accuracy on {\TASKNAME}-500 reduces to near-chance (dashed lines) for all models as grammar size (number of nonlexical productions in the grammar, \textbf{top row}) and string length (\textbf{bottom row}) increase.}
    \label{fig:accuracy_by_complexity}
\end{figure*}

\subsection{Performance Decreases as Grammar and Example Complexity Increases}
\label{sec:results-properties}

For all models, performance on {\TASKNAME}-500 decreases as a function of a grammar's complexity, as quantified by each of the four grammar parameters. Among these parameters, the number of nonlexical productions $\Nnonlex$ is most strongly anti-correlated with performance. On a per-model basis, we observe a roughly log-linear relationship between the number of nonlexical productions and performance: though some models have high accuracy on small grammars, as $\Nnonlex$ approaches 500 all models are at or below chance performance (\cref{fig:accuracy_by_complexity}, top). 

Model performance is also affected by the complexity of individual examples. As example length $\ell$ increases, models' mean accuracy over both positive and negative examples decreases (\cref{fig:accuracy_by_complexity}, bottom). This drop-off happens quite rapidly, with an inflection point occurring between $\ell = 5$ and $\ell = 15$ depending on the model. A regression shows that the effects of $\log(\Nnonlex)$ and $\log(\ell)$ are highly significant (see~\cref{tab:corr,tab:regression} in~\cref{sec:grammar-stats}).

\subsection{Models Generally Agree on Which Grammars and Examples are Hard}

\begin{figure*}
    \centering
    \includegraphics[width=\linewidth]{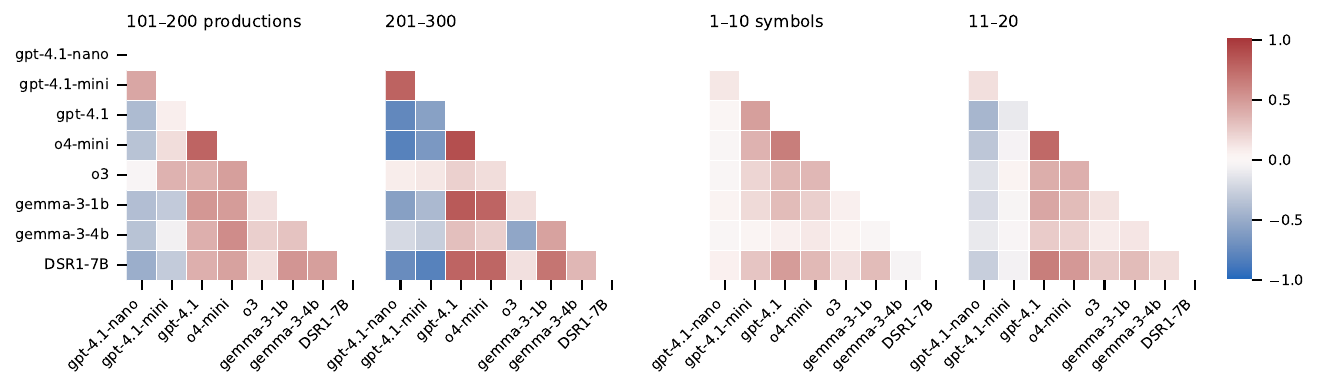}
    \caption{Spearman's rank correlation coefficients for the mean accuracy per grammar (\textbf{left}) and per example (\textbf{right}) between different models, binned by complexity. Most models agree on which grammars and examples are hard, with the exception of \texttt{gpt-4.1-nano/mini} which tend to disagree with other models. The strength of these correlations increases modestly as grammar and example complexity grow.}
    \label{fig:rank-correlations}
\end{figure*}

Alongside the strong correlation between accuracy and grammar complexity, we found substantial variability in accuracy among grammars of similar complexity; this is particularly the case for the models with higher overall accuracy, \texttt{gpt-4.1-mini}, \texttt{o4-mini} and \texttt{o3} (\cref{fig:accuracy_by_complexity}). Are the same grammars or examples challenging for all models? To test this, we rank the grammars and examples by their accuracy for each model and compute the Spearman's rank correlation across models. To mitigate the overall correlation with grammar complexity, we first divide the grammars into five bins of 100 production rules each; we similarly divide examples into length bins, in increments of 10. We find that grammars (\cref{fig:rank-correlations}, left) and samples (\cref{fig:rank-correlations}, right) have generally consistent difficulty across models. Of note is the fact that two models, \texttt{gpt-4.1-nano} and \texttt{gpt-4.1-mini}, are less correlated with the remaining models but have positive correlation with one another; this behavior likely reflects the fact that these models have a bias in favor of predicting samples to be positive, unlike the pattern of all other models as shown in~\cref{fig:acc-by-type-length}. We also observe that these (anti-)correlations strengthen as grammar and example complexity increase: correlations are closer to zero on grammars with 100--200 productions than they are for those with 200--300 productions, and similarly for examples of lengths 1--10 and 11--20. For a full overview of these correlations, see~\cref{fig:sample-rank} in~\cref{sec:results-tables}.

\section{How Do LLMs Succeed and Fail?}

To understand how models attempt to solve RELIC, where they succeed, and why they fail, we examine the chains-of-thought produced by models prior to providing their final answers; we have access to intermediate outputs for all models except \texttt{o3} and \texttt{o4-mini}, whose CoTs are not shared with the user. We perform a qualitative analysis of model traces to identify high-level strategies models employ when responding. We then quantify how model strategy changes as a function of task complexity, and correlate these observations to changes in test-time compute expenditure.

We note that while theoretical \citep{merrill-2023-parallelism} and empirical \citep{turpin-2023-language-a,pfau-2024-letS} work shows that CoTs may not be faithful representatives of underlying computation, such separation arises mostly in adversarial or otherwise-biased settings, where LLMs have been deliberately trained or inadvertently prompted to elicit unfaithful CoTs; \citet{lanham-2023-measuring} found that such problems do not often arise in non-adversarial settings. As our setup does not contain any means to introduce such bias (zero-shot prompts preclude introducing spurious correlations in the exemplars), we suspect that the risk of unfaithfulness here is low.

\subsection{Qualitative Analysis of Strategy}
We observe that models can identify a valid strategy to solve RELIC: using the rules of the grammar to construct a parse tree of the string, or showing (by exhaustion or proof) that one cannot exist. In some cases, with small grammars and short strings, models can often follow that strategy through to arrive at the correct answer. In other cases, models fail because they get distracted by irrelevant information, hallucinate success criteria, or second-guess their (correct) choice of strategy and attempt to solve the problem by other, invalid means. 

\paragraph{Successful attempts.} In some cases, models correctly identify positive samples by identifying the relevant production rules and deriving a parse tree. The response below highlights such an instance in which \texttt{gpt-4.1-mini} is able to exhaustively search for derivations compatible with the provided example; this search involves recursively applying the rule \texttt{NT5 $\to$ NT0 NT5} three times, resulting in a full derivation for the 5-symbol string \texttt{t30 t24 t24 t23 t4}. The model also successfully ignores production rules irrelevant to the particular example.

\begin{llm}
Let's analyze the grammar carefully and the string to verify if the string can be generated by the grammar.

\llmfiller
\textbf{Step 1:} Since S -> NT5 NT2\\
The string breaks into two substrings:\\
- substring generated by NT5\\
- substring generated by NT2\\
Let's try possible splits:\\
- split after 1 token: (t30) (t24 t24 t23 t4)\\
- split after 2 tokens: (t30 t24) (t24 t23 t4)\\
- split after 3 tokens: (t30 t24 t24) (t23 t4)\\
- split after 4 tokens: (t30 t24 t24 t23) (t4)

Try each possibility.

\textbf{Step 2:} Can NT5 generate the initial substring?\\
\emph{NT5 options:}\\
- NT5 -> NT0 NT5  \\
- NT5 -> terminals like `t18', `t23', `t25', `t13' (no 't30' here)\\
\emph{NT0 terminals:}
\llmfiller
So parse:\\
- NT5:  \\
NT5 -> NT0 NT5  \\
NT0 -> `t30'  \\
NT5 -> NT0 NT5  \\
NT0 -> `t24'  \\
NT5 -> NT0 NT5  \\
NT0 -> `t24'  \\
NT5 -> `t23' (terminal)

Tokens: t30 t24 t24 t23

So yes, this derives the 4 tokens.

\textbf{Step 3:} Now check if NT2 can generate last token: `t4'\\
NT2 terminals: `t9', `t23', `t27', `t4', `t30'\\
Yes, `t4' is a terminal allowed by NT2 \\
Therefore, NT2 can directly generate `t4'

\textbf{Step 4:} Full derivation\\
S -> NT5 NT2\\
NT5 -> `t30 t24 t24 t23' (via recursive NT5 -> NT0 NT5)\\  
NT2 -> `t4' terminal

The full string `t30 t24 t24 t23 t4` is generated by S.

\textbf{Final conclusion:} The string can be generated by the grammar.

\textbf{Answer:} Yes
\end{llm}

\vspace*{0em}

\begin{figure*}[t]
    \centering
    \includegraphics[width=\linewidth]{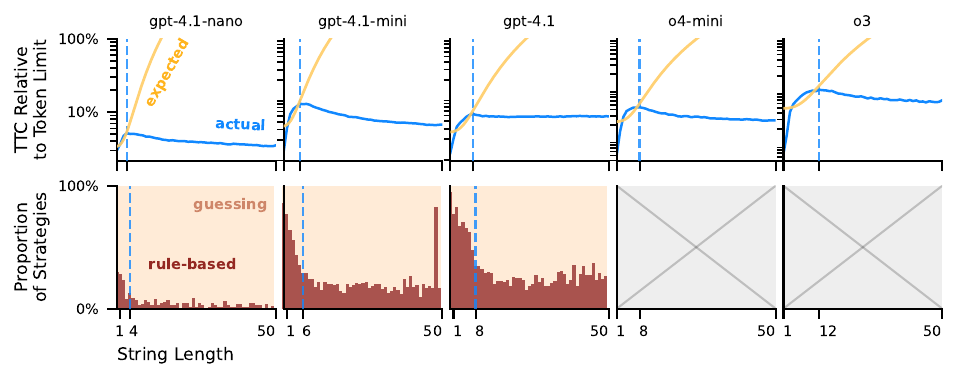}
    \caption{As string length increases, the test-time compute (TTC) expended by models peaks early and then diminishes (\textbf{top row}, blue line); this contrasts with the cubic growth that would be expected from applying the correct algorithm (yellow line). The cubic curve is fit to points before the peak, and TTC is computed as the mean number of completion tokens produced for examples of a given length, relative to the model's maximum token limit. As relative TTC stops increasing, models shift from pursuing rule-based approaches to guessing (\textbf{bottom row}, based on \texttt{gpt-5}'s classifications of the models' chains-of-thought). Neither \texttt{o4-mini} nor \texttt{o3} provide full CoTs, so we cannot classify their strategies directly, but the similarity in compute expenditure to models that report CoTs suggests a similar shift in strategy.}
    \label{fig:gpt-strategy}
\end{figure*}

\paragraph{Failed attempts.} Models often fail on more complicated versions of the task even when they are able to solve it in the simpler cases. These failures have many different causes. In some instances, models fail because they \emph{accumulate errors}. \Cref{sec:dsr1-20250218222557-a,sec:dsr1-20250402155408_676876-a} highlight such cases in which models describe a valid way to solve the problem but make invalid inferences from intermediate conclusions, become distracted by irrelevant information, or otherwise commit mistakes in the process of trying to follow a good strategy to its conclusion.

In other instances, models fail because they rely on \emph{invalid strategies}. Although the nature of the problem is unchanged, prompting models with more complex grammars and strings can cause models to move away from valid, rule-based approaches and instead attempt to solve the task based on irrelevant or invalid heuristics, such as string length or the number of different ways to derive a single terminal symbol. Importantly, these heuristics can partially coincide with correct conclusions, leading to models being right for the wrong reasons \citep{mccoy-2019-wrong}. \Cref{sec:dsr1-20250402155408_676876-b} shows just such a case, where a model rejects a string because it does not contain a particular symbol; while the response is correct, the reasoning is doubly invalid: not only does accepting the string not involve ensuring the presence of this symbol, but the symbol in question is not even in the grammar, and so its presence would be sufficient to reject the string.

\subsection{Quantitative Analysis of Strategy}
\label{sec:model-strategies}

In light of the finding that models can propose valid solutions to RELIC but don't always implement those solutions, we perform a quantitative analysis to explore how model strategy changes as a function of task complexity. We employ the ``LLM-as-a-judge'' framework \citep{zheng-2023-judging}, prompting \texttt{gpt-5} to classify model completions as either \emph{rule-based} (requiring a full derivation of the string, or a proof that one cannot exist) or \emph{heuristic} (appealing to properties of the example or grammar to argue why an example seems likely or unlikely to be derivable). We focus on the \texttt{gpt-4.1} series of models, since they can solve the task in simple cases, fail on complex ones, and provide full access to their CoTs. We perform human and LLM validation of the LLM-judge; see~\cref{sec:llm-judge} for details.

\Cref{fig:gpt-strategy} (bottom) shows that model strategy is strongly dependent on task complexity. When plotted against string length, the proportion of responses employing a rule-based approach peaks very early and then falls off quickly as all three models in the \texttt{gpt-4.1} series shift to using invalid heuristics to arrive at an answer.

\subsection{Test-Time Compute Grows Insufficiently} \label{sec:ttc-length} 

Models can scale their test-time compute (TTC) expenditure by producing more intermediate CoT tokens in a bid to solve harder problems. We examine how models expend TTC on RELIC by exploring how the length of the candidate strings impacts the number of tokens a model produces during inference (relative to the maximum token limit set for the model). Solving RELIC requires transformer LLMs to increase compute expenditure superlinearly in string length (\cref{sec:complexity-theory}). Yet \cref{fig:gpt-strategy} (top) shows that relative TTC grows \emph{sublinearly}, peaks early at levels far below ($\sim10\%$) the token limits set for each model, and then \emph{decreases} as a function of string length. Moreover, the point at which TTC peaks and starts to drop off occurs close to the peak and drop-off in preference for rule-based strategies over heuristic guesses, further supporting the hypothesis that the models stop attempting to apply the correct strategy once the string reaches a certain length.

\section{Finetuning on RELIC Does Not Improve Performance} \label{sec:ft-main}

\begin{figure}[ht!]
    \centering
    \includegraphics[width=\linewidth]{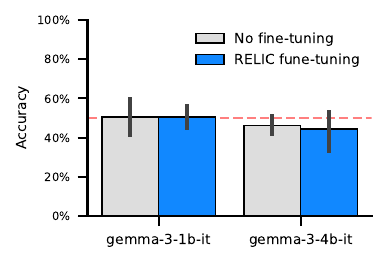}
    \caption{Finetuning the Gemma 3 models directly on RELIC grammar--string--label data does not confer any statistically significant benefit for the 1B or 4B models; error bars report 95\% confidence intervals for mean model accuracy over five random seeds.}
    \label{fig:gemma-ft}
\end{figure}

Could the weaker performance of smaller, open-weight models we study be due to a limited ability to infer the task from the prompt, as opposed to weak reasoning capabilities? To test this possibility, we directly tune the two Gemma 3 variants studied here (\texttt{-1b-it} and \texttt{-4b-it}) on a subset of RELIC-500. Since these two models are fine-tuned to be chat models, we format the fine-tuning samples using the Gemma chat templates for multi-turn conversations, with the \texttt{user} turn being the standard RELIC prompt used during evaluation and the \texttt{model} turn being the correct answer. See \cref{sec:ft} for details on fine-tuning.

\Cref{fig:gemma-ft} shows that directly fine-tuning the Gemma 3 models on grammar--string--label data has no statistically significant impact on model performance: for both the 1B and 4B models, mean validation accuracy after fine-tuning remained within the 95\% confidence intervals of the mean accuracies of the non-fine-tuned models. This suggests that these models' poor performance is not primarily due to inability to infer the task that needs to be performed from a single prompt.

\section{Discussion}
\label{sec:discussion}

\subsection{LLMs Quiet Quit on Hard Tasks}

Across a wide variety of contemporary LLMs, models often demonstrate a good high-level understanding of the RELIC task: models can correctly outline sound strategies to solve in-context language recognition and apply these strategies on simple instances of the task. Yet increasing the difficulty of RELIC by making strings longer and grammars larger---though still considerably smaller than the grammars of natural languages---induces a significant degradation in performance, even among the most capable reasoning models trained to ``think longer'' for harder problems. Quantitative analysis of their token expenditures shows that models stop increasing their inference compute expenditure as a function of task complexity quite early on, in contrast to theoretical requirements for maintaining accuracy on RELIC. Instead, compute expenditure peaks early, well short of the models' allotted compute budgets, and then \emph{decreases}. Qualitative analysis on models where reasoning traces are available reveals that this shift in compute expenditure accompanies a dramatic reversal in reasoning \emph{strategy:} as problems become harder, models are increasingly less inclined to apply sound reasoning strategies, instead opting to rely on shallow guesses.

We observe the same trend of diminishing TTC expenditure as a function of task complexity in the reasoning models for which we do not have access to reasoning traces (here, \texttt{o4-mini} and \texttt{o3}) as we do in those with visible intermediate reasoning (e.g., the \texttt{gpt-4.1} series). Since the computational complexity of RELIC requires that models grow their compute expenditure as a function of task difficulty, the observed behavior means that these reasoning models must be exhibiting the same shift in reasoning strategy which can be directly observed in the other models: as the task becomes harder, they start to guess. This results in a peculiar phenomenon where the best reasoning models must be ``quiet quitting'' on hard tasks which they could succeed at if they applied better strategies.

\subsection{\TASKNAME\ Predicts Reasoning Performance}

\begin{figure}
    \centering
    \includegraphics[width=\linewidth]{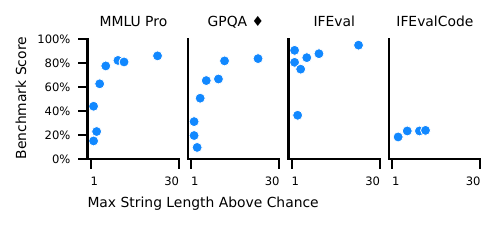}
    \caption{Performance on RELIC (here, measured by the longest string lengths for which models are reliably above their minimum accuracy, which for most models is at-chance) generally correlates with reasoning and instruction following evaluations.}
    \label{fig:benchmarks}
\end{figure}

CFL recognition may be hard for LLMs not only because of its computational complexity (\cref{sec:complexity-theory}) and context use requirements (\cref{sec:context-use}), but also because the task itself is out-of-distribution and involves using data presented in a particular format that may be under-represented in training. To validate that success on RELIC is not merely an artifact of the LLMs' inability to infer the task itself, we compare performance on RELIC to that on a variety of other benchmarks: reasoning and knowledge benchmarks MMLU Pro \citep{wang-2024-mmlu-pro} and GPQA Diamond \citep{rein-2024-gpqa}; IFEval \citep{zhou-2023-instruction-following}, an instruction following benchmark; and IFEvalCode \citep{yang-2025-ifevalcode}, an instruction-following benchmark specifically focusing on code generation. \Cref{fig:benchmarks} shows that performance on RELIC (as measured by the maximum length of strings that models can reliably classify above-chance) broadly correlates with performance on these more general-purpose benchmarks, suggesting that success or failure on RELIC is not merely an artifact of a model's ability to understand the task formatting but is more broadly correlated with its general-purpose utility on reasoning tasks.

\subsection{Scalably-Complex Tasks Can Evaluate Hidden Reasoning}

Proprietary reasoning models, such as \texttt{o4-mini} and \texttt{o3}, commonly obscure their reasoning traces, presenting the user with only their final response and potentially a summary of the hidden traces, along with metadata about how many tokens of compute were used to arrive at that final answer. The fact that these traces are withheld poses challenges for evaluation of such models, since it is impossible to conduct any qualitative analysis on the responses to determine whether the reasoning strategies employed are sound or consistent with a user's guidance.

Scalably-complex tasks like RELIC offer a partial window into the reasoning traces of these proprietary models. By connecting the intrinsic computational complexity of the task, here measured by the time-complexity of the best algorithms for solving RELIC as a function of its generating hyperparameters, to the computational behavior of language models, we can derive predictions for how test-time compute budgets \emph{must} increase as a function of task complexity; comparing this predicted behavior to what is observed empirically reveals whether closed-reasoning models are, in general, applying sound reasoning strategies.

\subsection{Models Struggle at Dense Context Use}

Models' poor performance on RELIC highlights that difficulties in context use can arise from a prompt's \emph{density} in addition to its length. The grammars and strings used in RELIC are small compared to the models' context windows, amounting to no more than a few hundred unary or binary production rules and strings of no more than 50 words. Despite this, the highly-compositional nature of the task presents challenges for current LLMs: since it is not \emph{prima facie} obvious which nonterminal rules must be retrieved from the context, nor the order in which they must be combined, RELIC requires models to effectively search over combinations of pieces of context to arrive at an answer, making the task much harder than many long-context use evaluations such as multi-hop needles or state evaluation benchmarks like NeedleBench \citep{li-2025-needlebench} or Michelangelo \citep{vodrahalli-2024-michelangelo}.

\subsection{Limitations}
\label{sec:limitations}

\paragraph{RELIC does not control for task familiarity.}

While RELIC's nature as a synthetic benchmark means we can be very confident that any particular evaluation sample is novel for an LLM, it is quite possible that different models have been exposed to the concept of CFL recognition and the associated solutions. As such, success on RELIC likely depends on models having some prior understanding of the problem setup and does not constitute evidence that models are able to develop novel \emph{solutions} to the task, but merely that they may be able to \emph{apply} known solutions to novel instances of the task. 

\paragraph{Strategy classification is subject to error.}
While our LLM-judge attained reasonable levels of agreement with human annotators on whether models were guessing versus attempting to employ good reasoning strategies when solving RELIC, the ability to further verify whether or not models make low-level mistakes in their chains of thought is not here addressed at scale. While we do observe instances of this, such as in the response shown in \cref{sec:dsr1-20250402155408_676876-b} where a model hallucinates a success condition based on a terminal symbol which doesn't appear in the grammar, we do not know how often such cases appear. Since these cases are difficult to automatically detect and verify, we think it possible that the LLM-judge strategy classifications may be overly rosy, as many instances classified by the LLMs as displaying rule-based reasoning may contain errors which make the arguments invalid or unsound.

\subsection{Future Directions}
\label{sec:future-work}

The results here suggest several avenues of future research. First, the negative results of our fine-tuning experiments, albeit on small models, imply that different strategies must be pursued to improve model performance. In particular, it may be the case that training models on more structured completions involving guided reasoning strategies, such as full derivations or exhaustive searches, is more effective than merely tuning on labeled outputs \citep{lambert-2025-tulu}. Likewise, using in-context samples to generalize to novel grammars may be a promising way to attain better fidelity to rule-based approaches. Explicit guidance on which particular strategies are to be used may further boost performance, though at the cost of requiring the user to do upfront work to identify what those solutions are. Taken as a proxy for more general reasoning tasks, it is more desirable to find ways of improving performance which do not require the user to proactively identify valid completion strategies on a per-prompt basis.

Second, extending the class of grammars surveyed here to include more complex languages, such as context-sensitive languages, may yield further insight into how LLMs fare on tasks that are computationally harder than what is tested here; extension to context-sensitive languages would provide a clear analogy to the task of determining the syntactic validity of programs in many programming languages, which have non-context free components such as macros in C.

Finally, it may be possible to leverage a similar experimental setup to test for LLMs' ability to reason about formal languages in tasks other than language recognition; a promising candidate is to explore whether or not reasoning models can use formal grammars as a means of \emph{translating} between two formal languages when given their specifications. Such an evaluation would provide a way to estimate how well LLMs can do in-context translation of natural languages \citep{tanzer-2023-benchmark-a,petty-2026-evaluating}.

\subsection*{Acknowledgments}

Thanks to the NYU Computation and Psycholinguistics Lab for discussion, and to the anonymous reviewers for their valuable feedback. This work was supported in part through the NYU IT High Performance Computing resources, services, and staff expertise. This project is supported by the National Science Foundation (NSF) under grant NRT-HDR: FUTURE as well as Grant No. IIS-2239862.

\bibliography{references}

@ARTICLE{li-2025-needlebench,
  title    = "{NeedleBench: Evaluating LLM Retrieval and Reasoning Across
              Varying Information Densities}",
  author   = "Li, Mo and Zhang, Songyang and Zhang, Taolin and Duan, Haodong and
              Liu, Yunxin and Chen, Kai",
  journal  = "Transactions on Machine Learning Research",
  year     =  2025,
  url      = "https://openreview.net/forum?id=cEvmIKsRw0",
  issn     = "2835-8856"
}

@MISC{petty-2026-evaluating,
  title         = "{Evaluating in-context translation with synchronous
                   context-free grammar transduction}",
  author        = "Petty, Jackson and Goe, Jaulie and Linzen, Tal",
  month         =  "8~" # apr,
  year          =  2026,
  url           = "http://arxiv.org/abs/2604.07320",
  archivePrefix = "arXiv",
  primaryClass  = "cs.CL",
  eprint        = "2604.07320",
  doi           = "10.48550/arXiv.2604.07320"
}

@MISC{vonWerra-2020-trl,
  title    = "{TRL: Transformers Reinforcement Learning}",
  author   = "von Werra, Leandro and Belkada, Younes and Tunstall, Lewis and
              Beeching, Edward and Thrush, Tristan and Lambert, Nathan and
              Huang, Shengyi and Rasul, Kashif and Gallouédec, Quentin",
  year     =  2020,
  url      = "https://github.com/huggingface/trl"
}

@INPROCEEDINGS{wolf-2020-transformers,
  title     = "{Transformers: State-of-the-art natural language processing}",
  author    = "Wolf, Thomas and Debut, Lysandre and Sanh, Victor and Chaumond,
               Julien and Delangue, Clement and Moi, Anthony and Cistac, Pierric
               and Rault, Tim and Louf, Remi and Funtowicz, Morgan and Davison,
               Joe and Shleifer, Sam and von Platen, Patrick and Ma, Clara and
               Jernite, Yacine and Plu, Julien and Xu, Canwen and Le Scao, Teven
               and Gugger, Sylvain and Drame, Mariama and Lhoest, Quentin and
               Rush, Alexander",
  booktitle = "{Proceedings of the 2020 Conference on Empirical Methods in
               Natural Language Processing: System Demonstrations}",
  publisher = "Association for Computational Linguistics",
  address   = "Stroudsburg, PA, USA",
  pages     = "38--45",
  month     =  oct,
  year      =  2020,
  url       = "http://dx.doi.org/10.18653/v1/2020.emnlp-demos.6",
  doi       = "10.18653/v1/2020.emnlp-demos.6"
}

@MISC{openai-2025-introducing,
  title        = "{Introducing GPT-4.1 in the {API}}",
  author       = "{OpenAI}",
  month        =  "15~" # apr,
  year         =  2025,
  howpublished = "\url{https://openai.com/index/gpt-4-1/}",
  note         = "Accessed: 2025-5-12",
  language     = "en"
}

@MISC{openai-2025-openai,
  title        = "{OpenAI o3 and o4-mini System Card}",
  author       = "{OpenAI}",
  month        =  "16~" # may,
  year         =  2025,
  howpublished = "\url{https://openai.com/index/o3-o4-mini-system-card/}",
  note         = "Accessed: 2025-5-12",
  language     = "en"
}

@MISC{vodrahalli-2024-michelangelo,
  title         = "{Michelangelo: Long context evaluations beyond haystacks via
                   Latent Structure Queries}",
  author        = "Vodrahalli, Kiran and Ontanon, Santiago and Tripuraneni,
                   Nilesh and Xu, Kelvin and Jain, Sanil and Shivanna, Rakesh
                   and Hui, Jeffrey and Dikkala, Nishanth and Kazemi, Mehran and
                   Fatemi, Bahare and Anil, Rohan and Dyer, Ethan and Shakeri,
                   Siamak and Vij, Roopali and Mehta, Harsh and Ramasesh, Vinay
                   and Le, Quoc and Chi, Ed and Lu, Yifeng and Firat, Orhan and
                   Lazaridou, Angeliki and Lespiau, Jean-Baptiste and Attaluri,
                   Nithya and Olszewska, Kate",
  month         =  "19~" # sep,
  year          =  2024,
  url           = "http://arxiv.org/abs/2409.12640",
  archivePrefix = "arXiv",
  primaryClass  = "cs.CL",
  eprint        = "2409.12640"
}

@MISC{bohnet-2024-long-span,
  title         = "{Long-span question-answering: Automatic question generation
                   and QA-system ranking via side-by-side evaluation}",
  author        = "Bohnet, Bernd and Swersky, Kevin and Liu, Rosanne and
                   Awasthi, Pranjal and Nova, Azade and Snaider, Javier and
                   Sedghi, Hanie and Parisi, Aaron T and Collins, Michael and
                   Lazaridou, Angeliki and Firat, Orhan and Fiedel, Noah",
  month         =  "31~" # may,
  year          =  2024,
  url           = "http://arxiv.org/abs/2406.00179",
  archivePrefix = "arXiv",
  primaryClass  = "cs.CL",
  eprint        = "2406.00179"
}

@INPROCEEDINGS{loshchilov-2019-decoupled,
  title     = "{Decoupled Weight Decay Regularization}",
  author    = "Loshchilov, Ilya and Hutter, Frank",
  booktitle = "{International Conference on Learning Representations}",
  year      =  2019,
  url       = "https://openreview.net/forum?id=Bkg6RiCqY7"
}

@MISC{openai-2024-learning,
  title        = "{Learning to reason with {LLMs}}",
  author       = "{OpenAI}",
  month        =  "12~" # sep,
  year         =  2024,
  howpublished = "\url{https://openai.com/index/learning-to-reason-with-llms/}",
  note         = "Accessed: 2025-5-9",
  language     = "en"
}

@INCOLLECTION{chomsky-1963-algebraic,
  title     = "{The Algebraic Theory of Context-Free Languages}",
  author    = "Chomsky, N and Schützenberger, M P",
  editor    = "Braffort, P and Hirschberg, D",
  booktitle = "{Studies in Logic and the Foundations of Mathematics}",
  publisher = "Elsevier",
  volume    =  35,
  pages     = "118--161",
  series    = "Computer Programming and Formal Systems",
  month     =  "1~" # jan,
  year      =  1963,
  url       = "https://www.sciencedirect.com/science/article/pii/S0049237X08720238",
  language  = "en"
}

@INPROCEEDINGS{kim-2020-cogs,
  title     = "{COGS: A Compositional Generalization Challenge Based on Semantic
               Interpretation}",
  author    = "Kim, Najoung and Linzen, Tal",
  booktitle = "{Proceedings of the 2020 Conference on Empirical Methods in
               Natural Language Processing (EMNLP)}",
  publisher = "Association for Computational Linguistics",
  address   = "Online",
  pages     = "9087–9105",
  month     =  nov,
  year      =  2020,
  url       = "https://aclanthology.org/2020.emnlp-main.731",
  doi       = "10.18653/v1/2020.emnlp-main.731"
}

@INPROCEEDINGS{clark-2017-testing,
  title     = "{Testing distributional properties of context-free grammars}",
  author    = "Clark, Alexander",
  editor    = "Verwer, Sicco and van Zaanen, Menno and Smetsers, Rick",
  booktitle = "{Proceedings of The 13th International Conference on Grammatical
               Inference}",
  publisher = "PMLR",
  address   = "Delft, The Netherlands",
  volume    =  57,
  pages     = "42--53",
  series    = "Proceedings of Machine Learning Research",
  year      =  2017,
  url       = "https://proceedings.mlr.press/v57/clark16.pdf"
}

@INPROCEEDINGS{wei-2022-chain-of-thought-a,
  title     = "{Chain-of-Thought Prompting Elicits Reasoning in Large Language
               Models}",
  author    = "Wei, Jason and Wang, Xuezhi and Schuurmans, Dale and Bosma,
               Maarten and Ichter, Brian and Xia, Fei and Chi, Ed H and Le, Quoc
               V and Zhou, Denny",
  booktitle = "{Advances in Neural Information Processing Systems}",
  month     =  "16~" # may,
  year      =  2022,
  url       = "https://openreview.net/pdf?id=_VjQlMeSB_J"
}

@INPROCEEDINGS{rein-2024-gpqa,
  title     = "{GPQA: A Graduate-Level Google-Proof Q\&A Benchmark}",
  author    = "Rein, David and Hou, Betty Li and Stickland, Asa Cooper and
               Petty, Jackson and Pang, Richard Yuanzhe and Dirani, Julien and
               Michael, Julian and Bowman, Samuel R",
  booktitle = "{First Conference on Language Modeling}",
  month     =  oct,
  year      =  2024,
  url       = "https://arxiv.org/abs/2311.12022"
}

@INPROCEEDINGS{dziri-2023-faith,
  title     = "{Faith and Fate: Limits of Transformers on Compositionality}",
  author    = "Dziri, Nouha and Lu, Ximing and Sclar, Melanie and Li, Xiang
               Lorraine and Jiang, Liwei and Lin, Bill Yuchen and Welleck, Sean
               and West, Peter and Bhagavatula, Chandra and Bras, Ronan Le and
               Hwang, Jena D and Sanyal, Soumya and Ren, Xiang and Ettinger,
               Allyson and Harchaoui, Zaid and Choi, Yejin",
  booktitle = "{Thirty-seventh Conference on Neural Information Processing
               Systems}",
  year      =  2023,
  url       = "https://openreview.net/forum?id=Fkckkr3ya8"
}

@INPROCEEDINGS{sanh-2022-multitask,
  title     = "{Multitask Prompted Training Enables Zero-Shot Task
               Generalization}",
  author    = "Sanh, Victor and Webson, Albert and Raffel, Colin and Bach,
               Stephen H and Sutawika, Lintang and Alyafeai, Zaid and Chaffin,
               Antoine and Stiegler, Arnaud and Raja, Arun and Dey, Manan and
               Bari, M Saiful and Xu, Canwen and Thakker, Urmish and Sharma,
               Shanya Sharma and Szczechla, Eliza and Kim, Taewoon and
               Chhablani, Gunjan and Nayak, Nihal V and Datta, Debajyoti and
               Chang, Jonathan and Jiang, Mike Tian-Jian and Wang, Han and
               Manica, Matteo and Shen, Sheng and Yong, Zheng Xin and Pandey,
               Harshit and Bawden, Rachel and Wang, Thomas and Neeraj, Trishala
               and Rozen, Jos and Sharma, Abheesht and Santilli, Andrea and
               Févry, Thibault and Fries, Jason Alan and Teehan, Ryan and Scao,
               Teven Le and Biderman, Stella and Gao, Leo and Wolf, Thomas and
               Rush, Alexander M",
  booktitle = "{The Tenth International Conference on Learning Representations,
               ICLR 2022, Virtual Event, April 25-29, 2022}",
  publisher = "OpenReview.net",
  year      =  2022,
  url       = "https://openreview.net/forum?id=9Vrb9D0WI4"
}

@ARTICLE{yao-2023-tree-a,
  title    = "{Tree of Thoughts: Deliberate Problem Solving with Large Language
              Models}",
  author   = "Yao, Shunyu and Yu, Dian and Zhao, Jeffrey and Shafran, Izhak and
              Griffiths, Tom and Cao, Yuan and Narasimhan, Karthik",
  journal  = "Advances in Neural Information Processing Systems",
  volume   =  36,
  pages    = "11809--11822",
  month    =  "15~" # dec,
  year     =  2023,
  url      = "https://proceedings.neurips.cc/paper/2023/hash/271db9922b8d1f4dd7aaef84ed5ac703-Abstract.html"
}

@MISC{kamradt-2023-gkamradt/llmtest_needleinahaystack,
  title        = "{gkamradt/{LLMTest\_NeedleInAHaystack}}",
  author       = "Kamradt, Greg",
  booktitle    = "{GitHub}",
  year         =  2023,
  howpublished = "\url{https://github.com/gkamradt/LLMTest\_NeedleInAHaystack/blob/main/README.md}",
  note         = "Accessed: 2025-4-10",
  language     = "en"
}

@MISC{clark-2018-syntheticpcfg,
  title       = "{syntheticpcfg: Code for generating synthetic PCFGs for testing
                 grammatical inference algorithms}",
  author      = "Clark, Alexander",
  institution = "Github",
  year        =  2018,
  url         = "https://github.com/alexc17/syntheticpcfg",
  language    = "en"
}

@ARTICLE{lee-2002-context-free,
  title     = "{Fast context-free grammar parsing requires fast boolean matrix
               multiplication}",
  author    = "Lee, Lillian",
  journal   = "Journal of the ACM",
  publisher = "Association for Computing Machinery (ACM)",
  volume    =  49,
  number    =  1,
  pages     = "1--15",
  month     =  "1~" # jan,
  year      =  2002,
  url       = "https://dl.acm.org/doi/10.1145/505241.505242",
  doi       = "10.1145/505241.505242",
  issn      = "0004-5411,1557-735X",
  language  = "en"
}

@ARTICLE{merrill-2023-parallelism,
  title     = "{The parallelism tradeoff: Limitations of log-precision
               transformers}",
  author    = "Merrill, William and Sabharwal, Ashish",
  journal   = "Transactions of the Association for Computational Linguistics",
  publisher = "MIT Press",
  volume    =  11,
  pages     = "531--545",
  month     =  "12~" # jun,
  year      =  2023,
  url       = "https://dx.doi.org/10.1162/tacl_a_00562",
  doi       = "10.1162/tacl\_a\_00562",
  issn      = "2307-387X",
  language  = "en"
}

@ARTICLE{ruzzo-1980-tree-size,
  title     = "{Tree-size bounded alternation}",
  author    = "Ruzzo, Walter L",
  journal   = "Journal of computer and system sciences",
  publisher = "Elsevier BV",
  volume    =  21,
  number    =  2,
  pages     = "218--235",
  month     =  "1~" # oct,
  year      =  1980,
  url       = "http://dx.doi.org/10.1016/0022-0000(80)90036-7",
  doi       = "10.1016/0022-0000(80)90036-7",
  issn      = "0022-0000,1090-2724",
  language  = "en"
}

@ARTICLE{venkateswaran-1991-properties,
  title     = "{Properties that characterize {LOGCFL}}",
  author    = "Venkateswaran, H",
  journal   = "Journal of computer and system sciences",
  publisher = "Elsevier BV",
  volume    =  43,
  number    =  2,
  pages     = "380--404",
  month     =  "1~" # oct,
  year      =  1991,
  url       = "http://dx.doi.org/10.1016/0022-0000(91)90020-6",
  doi       = "10.1016/0022-0000(91)90020-6",
  issn      = "0022-0000,1090-2724",
  language  = "en"
}

@TECHREPORT{greenlaw-1991-compendium,
  title       = "{A compendium of problems complete for P (preliminary)}",
  author      = "Greenlaw, Raymond and Ruzzo, Walter L and Hoover, James",
  publisher   = "University of Alberta Libraries",
  institution = "University of Alberta",
  year        =  1991,
  url         = "https://era.library.ualberta.ca/items/403292c5-460b-49e6-8b05-9a5a7b45b0d6",
  doi         = "10.7939/R39Z90F7X",
  language    = "en"
}

@INPROCEEDINGS{mccoy-2019-wrong,
  title     = "{Right for the wrong reasons: Diagnosing syntactic heuristics in
               natural language inference}",
  author    = "McCoy, Tom and Pavlick, Ellie and Linzen, Tal",
  booktitle = "{Proceedings of the 57th Annual Meeting of the Association for
               Computational Linguistics}",
  publisher = "Association for Computational Linguistics",
  address   = "Stroudsburg, PA, USA",
  pages     = "3428--3448",
  year      =  2019,
  url       = "http://dx.doi.org/10.18653/v1/P19-1334",
  doi       = "10.18653/v1/p19-1334"
}

@TECHREPORT{gemmaTeam-2025-gemma,
  title       = "{Gemma 3 Technical Report}",
  author      = "{Gemma Team} and Kamath, Aishwarya and Ferret, Johan and
                 Pathak, Shreya and Vieillard, Nino and Merhej, Ramona and
                 Perrin, Sarah and Matejovicova, Tatiana and Ramé, Alexandre and
                 Rivière, Morgane and Rouillard, Louis and Mesnard, Thomas and
                 Cideron, Geoffrey and Grill, Jean-Bastien and Ramos, Sabela and
                 Yvinec, Edouard and Casbon, Michelle and Pot, Etienne and
                 Penchev, Ivo and Liu, Gaël and Visin, Francesco and Kenealy,
                 Kathleen and Beyer, Lucas and Zhai, Xiaohai and Tsitsulin,
                 Anton and Busa-Fekete, Robert and Feng, Alex and Sachdeva,
                 Noveen and Coleman, Benjamin and Gao, Yi and Mustafa, Basil and
                 Barr, Iain and Parisotto, Emilio and Tian, David and Eyal,
                 Matan and Cherry, Colin and Peter, Jan-Thorsten and
                 Sinopalnikov, Danila and Bhupatiraju, Surya and Agarwal,
                 Rishabh and Kazemi, Mehran and Malkin, Dan and Kumar, Ravin and
                 Vilar, David and Brusilovsky, Idan and Luo, Jiaming and
                 Steiner, Andreas and Friesen, Abe and Sharma, Abhanshu and
                 Sharma, Abheesht and Gilady, Adi Mayrav and Goedeckemeyer,
                 Adrian and Saade, Alaa and Feng, Alex and Kolesnikov, Alexander
                 and Bendebury, Alexei and Abdagic, Alvin and Vadi, Amit and
                 György, András and Pinto, André Susano and Das, Anil and Bapna,
                 Ankur and Miech, Antoine and Yang, Antoine and Paterson,
                 Antonia and Shenoy, Ashish and Chakrabarti, Ayan and Piot,
                 Bilal and Wu, Bo and Shahriari, Bobak and Petrini, Bryce and
                 Chen, Charlie and Lan, Charline Le and Choquette-Choo,
                 Christopher A and Carey, C J and Brick, Cormac and Deutsch,
                 Daniel and Eisenbud, Danielle and Cattle, Dee and Cheng, Derek
                 and Paparas, Dimitris and Sreepathihalli, Divyashree Shivakumar
                 and Reid, Doug and Tran, Dustin and Zelle, Dustin and Noland,
                 Eric and Huizenga, Erwin and Kharitonov, Eugene and Liu,
                 Frederick and Amirkhanyan, Gagik and Cameron, Glenn and
                 Hashemi, Hadi and Klimczak-Plucińska, Hanna and Singh, Harman
                 and Mehta, Harsh and Lehri, Harshal Tushar and Hazimeh, Hussein
                 and Ballantyne, Ian and Szpektor, Idan and Nardini, Ivan and
                 Pouget-Abadie, Jean and Chan, Jetha and Stanton, Joe and
                 Wieting, John and Lai, Jonathan and Orbay, Jordi and Fernandez,
                 Joseph and Newlan, Josh and Ji, Ju-Yeong and Singh, Jyotinder
                 and Black, Kat and Yu, Kathy and Hui, Kevin and Vodrahalli,
                 Kiran and Greff, Klaus and Qiu, Linhai and Valentine, Marcella
                 and Coelho, Marina and Ritter, Marvin and Hoffman, Matt and
                 Watson, Matthew and Chaturvedi, Mayank and Moynihan, Michael
                 and Ma, Min and Babar, Nabila and Noy, Natasha and Byrd, Nathan
                 and Roy, Nick and Momchev, Nikola and Chauhan, Nilay and
                 Sachdeva, Noveen and Bunyan, Oskar and Botarda, Pankil and
                 Caron, Paul and Rubenstein, Paul Kishan and Culliton, Phil and
                 Schmid, Philipp and Sessa, Pier Giuseppe and Xu, Pingmei and
                 Stanczyk, Piotr and Tafti, Pouya and Shivanna, Rakesh and Wu,
                 Renjie and Pan, Renke and Rokni, Reza and Willoughby, Rob and
                 Vallu, Rohith and Mullins, Ryan and Jerome, Sammy and Smoot,
                 Sara and Girgin, Sertan and Iqbal, Shariq and Reddy, Shashir
                 and Sheth, Shruti and Põder, Siim and Bhatnagar, Sijal and
                 Panyam, Sindhu Raghuram and Eiger, Sivan and Zhang, Susan and
                 Liu, Tianqi and Yacovone, Trevor and Liechty, Tyler and Kalra,
                 Uday and Evci, Utku and Misra, Vedant and Roseberry, Vincent
                 and Feinberg, Vlad and Kolesnikov, Vlad and Han, Woohyun and
                 Kwon, Woosuk and Chen, Xi and Chow, Yinlam and Zhu, Yuvein and
                 Wei, Zichuan and Egyed, Zoltan and Cotruta, Victor and Giang,
                 Minh and Kirk, Phoebe and Rao, Anand and Black, Kat and Babar,
                 Nabila and Lo, Jessica and Moreira, Erica and Martins, Luiz
                 Gustavo and Sanseviero, Omar and Gonzalez, Lucas and Gleicher,
                 Zach and Warkentin, Tris and Mirrokni, Vahab and Senter, Evan
                 and Collins, Eli and Barral, Joelle and Ghahramani, Zoubin and
                 Hadsell, Raia and Matias, Yossi and Sculley, D and Petrov, Slav
                 and Fiedel, Noah and Shazeer, Noam and Vinyals, Oriol and Dean,
                 Jeff and Hassabis, Demis and Kavukcuoglu, Koray and Farabet,
                 Clement and Buchatskaya, Elena and Alayrac, Jean-Baptiste and
                 Anil, Rohan and {Dmitry} and {Lepikhin} and Borgeaud, Sebastian
                 and Bachem, Olivier and Joulin, Armand and Andreev, Alek and
                 Hardin, Cassidy and Dadashi, Robert and Hussenot, Léonard",
  institution = "Google DeepMind",
  month       =  "25~" # mar,
  year        =  2025
}

@TECHREPORT{deepseek-ai-2025-deepseek-r1,
  title       = "{DeepSeek-R1: Incentivizing Reasoning Capability in LLMs via
                 Reinforcement Learning}",
  author      = "{DeepSeek-AI} and Guo, Daya and Yang, Dejian and Zhang, Haowei
                 and Song, Junxiao and Zhang, Ruoyu and Xu, Runxin and Zhu,
                 Qihao and Ma, Shirong and Wang, Peiyi and Bi, Xiao and Zhang,
                 Xiaokang and Yu, Xingkai and Wu, Yu and Wu, Z F and Gou, Zhibin
                 and Shao, Zhihong and Li, Zhuoshu and Gao, Ziyi and Liu, Aixin
                 and Xue, Bing and Wang, Bingxuan and Wu, Bochao and Feng, Bei
                 and Lu, Chengda and Zhao, Chenggang and Deng, Chengqi and
                 Zhang, Chenyu and Ruan, Chong and Dai, Damai and Chen, Deli and
                 Ji, Dongjie and Li, Erhang and Lin, Fangyun and Dai, Fucong and
                 Luo, Fuli and Hao, Guangbo and Chen, Guanting and Li, Guowei
                 and Zhang, H and Bao, Han and Xu, Hanwei and Wang, Haocheng and
                 Ding, Honghui and Xin, Huajian and Gao, Huazuo and Qu, Hui and
                 Li, Hui and Guo, Jianzhong and Li, Jiashi and Wang, Jiawei and
                 Chen, Jingchang and Yuan, Jingyang and Qiu, Junjie and Li,
                 Junlong and Cai, J L and Ni, Jiaqi and Liang, Jian and Chen,
                 Jin and Dong, Kai and Hu, Kai and Gao, Kaige and Guan, Kang and
                 Huang, Kexin and Yu, Kuai and Wang, Lean and Zhang, Lecong and
                 Zhao, Liang and Wang, Litong and Zhang, Liyue and Xu, Lei and
                 Xia, Leyi and Zhang, Mingchuan and Zhang, Minghua and Tang,
                 Minghui and Li, Meng and Wang, Miaojun and Li, Mingming and
                 Tian, Ning and Huang, Panpan and Zhang, Peng and Wang,
                 Qiancheng and Chen, Qinyu and Du, Qiushi and Ge, Ruiqi and
                 Zhang, Ruisong and Pan, Ruizhe and Wang, Runji and Chen, R J
                 and Jin, R L and Chen, Ruyi and Lu, Shanghao and Zhou, Shangyan
                 and Chen, Shanhuang and Ye, Shengfeng and Wang, Shiyu and Yu,
                 Shuiping and Zhou, Shunfeng and Pan, Shuting and Li, S S and
                 Zhou, Shuang and Wu, Shaoqing and Ye, Shengfeng and Yun, Tao
                 and Pei, Tian and Sun, Tianyu and Wang, T and Zeng, Wangding
                 and Zhao, Wanjia and Liu, Wen and Liang, Wenfeng and Gao,
                 Wenjun and Yu, Wenqin and Zhang, Wentao and Xiao, W L and An,
                 Wei and Liu, Xiaodong and Wang, Xiaohan and Chen, Xiaokang and
                 Nie, Xiaotao and Cheng, Xin and Liu, Xin and Xie, Xin and Liu,
                 Xingchao and Yang, Xinyu and Li, Xinyuan and Su, Xuecheng and
                 Lin, Xuheng and Li, X Q and Jin, Xiangyue and Shen, Xiaojin and
                 Chen, Xiaosha and Sun, Xiaowen and Wang, Xiaoxiang and Song,
                 Xinnan and Zhou, Xinyi and Wang, Xianzu and Shan, Xinxia and
                 Li, Y K and Wang, Y Q and Wei, Y X and Zhang, Yang and Xu,
                 Yanhong and Li, Yao and Zhao, Yao and Sun, Yaofeng and Wang,
                 Yaohui and Yu, Yi and Zhang, Yichao and Shi, Yifan and Xiong,
                 Yiliang and He, Ying and Piao, Yishi and Wang, Yisong and Tan,
                 Yixuan and Ma, Yiyang and Liu, Yiyuan and Guo, Yongqiang and
                 Ou, Yuan and Wang, Yuduan and Gong, Yue and Zou, Yuheng and He,
                 Yujia and Xiong, Yunfan and Luo, Yuxiang and You, Yuxiang and
                 Liu, Yuxuan and Zhou, Yuyang and Zhu, Y X and Xu, Yanhong and
                 Huang, Yanping and Li, Yaohui and Zheng, Yi and Zhu, Yuchen and
                 Ma, Yunxian and Tang, Ying and Zha, Yukun and Yan, Yuting and
                 Ren, Z Z and Ren, Zehui and Sha, Zhangli and Fu, Zhe and Xu,
                 Zhean and Xie, Zhenda and Zhang, Zhengyan and Hao, Zhewen and
                 Ma, Zhicheng and Yan, Zhigang and Wu, Zhiyu and Gu, Zihui and
                 Zhu, Zijia and Liu, Zijun and Li, Zilin and Xie, Ziwei and
                 Song, Ziyang and Pan, Zizheng and Huang, Zhen and Xu, Zhipeng
                 and Zhang, Zhongyu and Zhang, Zhen",
  institution = "DeepSeek AI",
  month       =  "22~" # jan,
  year        =  2025
}

@MISC{gupta-2025-randomly,
  title         = "{Randomly sampled language reasoning problems reveal limits
                   of {LLMs}}",
  author        = "Gupta, Kavi and Sanders, Kate and Solar-Lezama, Armando",
  month         =  "6~" # jan,
  year          =  2025,
  url           = "http://arxiv.org/abs/2501.02825",
  archivePrefix = "arXiv",
  primaryClass  = "cs.LG",
  eprint        = "2501.02825"
}

@INPROCEEDINGS{deletang-2023-neural,
  title     = "{Neural Networks and the Chomsky Hierarchy}",
  author    = "Delétang, Grégoire and Ruoss, Anian and Grau-Moya, Jordi and
               Genewein, Tim and Wenliang, Li Kevin and Catt, Elliot and Cundy,
               Chris and Hutter, Marcus and Legg, Shane and Veness, Joel and
               Ortega, Pedro A",
  booktitle = "{The Eleventh International Conference on Learning
               Representations}",
  year      =  2023,
  url       = "https://openreview.net/forum?id=WbxHAzkeQcn"
}

@INPROCEEDINGS{wei-2022-finetuned,
  title     = "{Finetuned Language Models are Zero-Shot Learners}",
  author    = "Wei, Jason and Bosma, Maarten and Zhao, Vincent and Guu, Kelvin
               and Yu, Adams Wei and Lester, Brian and Du, Nan and Dai, Andrew M
               and Le, Quoc V",
  booktitle = "{International Conference on Learning Representations}",
  year      =  2022,
  url       = "https://openreview.net/forum?id=gEZrGCozdqR"
}

@INPROCEEDINGS{press-2023-measuring,
  title     = "{Measuring and narrowing the compositionality gap in language
               models}",
  author    = "Press, Ofir and Zhang, Muru and Min, Sewon and Schmidt, Ludwig
               and Smith, Noah and Lewis, Mike",
  booktitle = "{Findings of the Association for Computational Linguistics: EMNLP
               2023}",
  publisher = "Association for Computational Linguistics",
  address   = "Stroudsburg, PA, USA",
  pages     = "5687--5711",
  month     =  dec,
  year      =  2023,
  url       = "https://aclanthology.org/2023.findings-emnlp.378/",
  doi       = "10.18653/v1/2023.findings-emnlp.378"
}

@INPROCEEDINGS{zhou-2024-webarena,
  title     = "{WebArena: A Realistic Web Environment for Building Autonomous
               Agents}",
  author    = "Zhou, Shuyan and Xu, Frank F and Zhu, Hao and Zhou, Xuhui and Lo,
               Robert and Sridhar, Abishek and Cheng, Xianyi and Ou, Tianyue and
               Bisk, Yonatan and Fried, Daniel and Alon, Uri and Neubig, Graham",
  booktitle = "{The Twelfth International Conference on Learning
               Representations}",
  year      =  2024,
  url       = "https://openreview.net/forum?id=oKn9c6ytLx"
}

@INPROCEEDINGS{bhattamishra-2020-ability,
  title     = "{On the ability and limitations of transformers to recognize
               formal languages}",
  author    = "Bhattamishra, Satwik and Ahuja, Kabir and Goyal, Navin",
  booktitle = "{Proceedings of the 2020 Conference on Empirical Methods in
               Natural Language Processing (EMNLP)}",
  publisher = "Association for Computational Linguistics",
  address   = "Stroudsburg, PA, USA",
  pages     = "7096--7116",
  month     =  nov,
  year      =  2020,
  url       = "http://dx.doi.org/10.18653/v1/2020.emnlp-main.576",
  doi       = "10.18653/v1/2020.emnlp-main.576"
}

@ARTICLE{booth-1973-applying,
  title     = "{Applying probability measures to abstract languages}",
  author    = "Booth, T L and Thompson, R A",
  journal   = "IEEE transactions on computers. Institute of Electrical and
               Electronics Engineers",
  publisher = "Institute of Electrical and Electronics Engineers (IEEE)",
  volume    = "C-22",
  number    =  5,
  pages     = "442--450",
  month     =  may,
  year      =  1973,
  url       = "https://ieeexplore.ieee.org/document/1672339",
  doi       = "10.1109/t-c.1973.223746",
  issn      = "0018-9340",
  language  = "en"
}

@INPROCEEDINGS{zheng-2023-judging,
  title     = "{Judging LLM-as-a-Judge with MT-Bench and Chatbot Arena}",
  author    = "Zheng, Lianmin and Chiang, Wei-Lin and Sheng, Ying and Zhuang,
               Siyuan and Wu, Zhanghao and Zhuang, Yonghao and Lin, Zi and Li,
               Zhuohan and Li, Dacheng and Xing, Eric and Zhang, Hao and
               Gonzalez, Joseph E and Stoica, Ion",
  booktitle = "{Thirty-seventh Conference on Neural Information Processing
               Systems Datasets and Benchmarks Track}",
  year      =  2023,
  url       = "https://openreview.net/forum?id=uccHPGDlao"
}

@MISC{wu-2025-transformers,
  title         = "{How do transformers learn variable binding in symbolic
                   programs?}",
  author        = "Wu, Yiwei and Geiger, Atticus and Millière, Raphaël",
  month         =  "27~" # may,
  year          =  2025,
  url           = "http://arxiv.org/abs/2505.20896",
  archivePrefix = "arXiv",
  primaryClass  = "cs.LG",
  eprint        = "2505.20896",
  doi           = "10.48550/arXiv.2505.20896"
}

@MISC{openai-2024-openai,
  title         = "{OpenAI o1 System Card}",
  author        = "{OpenAI} and {:} and Jaech, Aaron and Kalai, Adam and Lerer,
                   Adam and Richardson, Adam and El-Kishky, Ahmed and Low, Aiden
                   and Helyar, Alec and Madry, Aleksander and Beutel, Alex and
                   Carney, Alex and Iftimie, Alex and Karpenko, Alex and Passos,
                   Alex Tachard and Neitz, Alexander and Prokofiev, Alexander
                   and Wei, Alexander and Tam, Allison and Bennett, Ally and
                   Kumar, Ananya and Saraiva, Andre and Vallone, Andrea and
                   Duberstein, Andrew and Kondrich, Andrew and Mishchenko,
                   Andrey and Applebaum, Andy and Jiang, Angela and Nair, Ashvin
                   and Zoph, Barret and Ghorbani, Behrooz and Rossen, Ben and
                   Sokolowsky, Benjamin and Barak, Boaz and McGrew, Bob and
                   Minaiev, Borys and Hao, Botao and Baker, Bowen and Houghton,
                   Brandon and McKinzie, Brandon and Eastman, Brydon and
                   Lugaresi, Camillo and Bassin, Cary and Hudson, Cary and Li,
                   Chak Ming and de Bourcy, Charles and Voss, Chelsea and Shen,
                   Chen and Zhang, Chong and Koch, Chris and Orsinger, Chris and
                   Hesse, Christopher and Fischer, Claudia and Chan, Clive and
                   Roberts, Dan and Kappler, Daniel and Levy, Daniel and Selsam,
                   Daniel and Dohan, David and Farhi, David and Mely, David and
                   Robinson, David and Tsipras, Dimitris and Li, Doug and
                   Oprica, Dragos and Freeman, Eben and Zhang, Eddie and Wong,
                   Edmund and Proehl, Elizabeth and Cheung, Enoch and Mitchell,
                   Eric and Wallace, Eric and Ritter, Erik and Mays, Evan and
                   Wang, Fan and Such, Felipe Petroski and Raso, Filippo and
                   Leoni, Florencia and Tsimpourlas, Foivos and Song, Francis
                   and von Lohmann, Fred and Sulit, Freddie and Salmon, Geoff
                   and Parascandolo, Giambattista and Chabot, Gildas and Zhao,
                   Grace and Brockman, Greg and Leclerc, Guillaume and Salman,
                   Hadi and Bao, Haiming and Sheng, Hao and Andrin, Hart and
                   Bagherinezhad, Hessam and Ren, Hongyu and Lightman, Hunter
                   and Chung, Hyung Won and Kivlichan, Ian and O'Connell, Ian
                   and Osband, Ian and Gilaberte, Ignasi Clavera and Akkaya,
                   Ilge and Kostrikov, Ilya and Sutskever, Ilya and Kofman,
                   Irina and Pachocki, Jakub and Lennon, James and Wei, Jason
                   and Harb, Jean and Twore, Jerry and Feng, Jiacheng and Yu,
                   Jiahui and Weng, Jiayi and Tang, Jie and Yu, Jieqi and
                   Candela, Joaquin Quiñonero and Palermo, Joe and Parish, Joel
                   and Heidecke, Johannes and Hallman, John and Rizzo, John and
                   Gordon, Jonathan and Uesato, Jonathan and Ward, Jonathan and
                   Huizinga, Joost and Wang, Julie and Chen, Kai and Xiao, Kai
                   and Singhal, Karan and Nguyen, Karina and Cobbe, Karl and
                   Shi, Katy and Wood, Kayla and Rimbach, Kendra and Gu-Lemberg,
                   Keren and Liu, Kevin and Lu, Kevin and Stone, Kevin and Yu,
                   Kevin and Ahmad, Lama and Yang, Lauren and Liu, Leo and
                   Maksin, Leon and Ho, Leyton and Fedus, Liam and Weng, Lilian
                   and Li, Linden and McCallum, Lindsay and Held, Lindsey and
                   Kuhn, Lorenz and Kondraciuk, Lukas and Kaiser, Lukasz and
                   Metz, Luke and Boyd, Madelaine and Trebacz, Maja and
                   Joglekar, Manas and Chen, Mark and Tintor, Marko and Meyer,
                   Mason and Jones, Matt and Kaufer, Matt and Schwarzer, Max and
                   Shah, Meghan and Yatbaz, Mehmet and Guan, Melody Y and Xu,
                   Mengyuan and Yan, Mengyuan and Glaese, Mia and Chen, Mianna
                   and Lampe, Michael and Malek, Michael and Wang, Michele and
                   Fradin, Michelle and McClay, Mike and Pavlov, Mikhail and
                   Wang, Miles and Wang, Mingxuan and Murati, Mira and Bavarian,
                   Mo and Rohaninejad, Mostafa and McAleese, Nat and Chowdhury,
                   Neil and Chowdhury, Neil and Ryder, Nick and Tezak, Nikolas
                   and Brown, Noam and Nachum, Ofir and Boiko, Oleg and Murk,
                   Oleg and Watkins, Olivia and Chao, Patrick and Ashbourne,
                   Paul and Izmailov, Pavel and Zhokhov, Peter and Dias, Rachel
                   and Arora, Rahul and Lin, Randall and Lopes, Rapha Gontijo
                   and Gaon, Raz and Miyara, Reah and Leike, Reimar and Hwang,
                   Renny and Garg, Rhythm and Brown, Robin and James, Roshan and
                   Shu, Rui and Cheu, Ryan and Greene, Ryan and Jain, Saachi and
                   Altman, Sam and Toizer, Sam and Toyer, Sam and Miserendino,
                   Samuel and Agarwal, Sandhini and Hernandez, Santiago and
                   Baker, Sasha and McKinney, Scott and Yan, Scottie and Zhao,
                   Shengjia and Hu, Shengli and Santurkar, Shibani and
                   Chaudhuri, Shraman Ray and Zhang, Shuyuan and Fu, Siyuan and
                   Papay, Spencer and Lin, Steph and Balaji, Suchir and Sanjeev,
                   Suvansh and Sidor, Szymon and Broda, Tal and Clark, Aidan and
                   Wang, Tao and Gordon, Taylor and Sanders, Ted and Patwardhan,
                   Tejal and Sottiaux, Thibault and Degry, Thomas and Dimson,
                   Thomas and Zheng, Tianhao and Garipov, Timur and Stasi, Tom
                   and Bansal, Trapit and Creech, Trevor and Peterson, Troy and
                   Eloundou, Tyna and Qi, Valerie and Kosaraju, Vineet and
                   Monaco, Vinnie and Pong, Vitchyr and Fomenko, Vlad and Zheng,
                   Weiyi and Zhou, Wenda and McCabe, Wes and Zaremba, Wojciech
                   and Dubois, Yann and Lu, Yinghai and Chen, Yining and Cha,
                   Young and Bai, Yu and He, Yuchen and Zhang, Yuchen and Wang,
                   Yunyun and Shao, Zheng and Li, Zhuohan",
  journal       = "arXiv [cs.AI]",
  year          =  2024,
  url           = "http://arxiv.org/abs/2412.16720",
  archivePrefix = "arXiv",
  primaryClass  = "cs.AI",
  eprint        = "2412.16720"
}

@MISC{lanham-2023-measuring,
  title         = "{Measuring faithfulness in Chain-of-Thought reasoning}",
  author        = "Lanham, Tamera and Chen, Anna and Radhakrishnan, Ansh and
                   Steiner, Benoit and Denison, Carson and Hernandez, Danny and
                   Li, Dustin and Durmus, Esin and Hubinger, Evan and Kernion,
                   Jackson and Lukošiūtė, Kamilė and Nguyen, Karina and Cheng,
                   Newton and Joseph, Nicholas and Schiefer, Nicholas and
                   Rausch, Oliver and Larson, Robin and McCandlish, Sam and
                   Kundu, Sandipan and Kadavath, Saurav and Yang, Shannon and
                   Henighan, Thomas and Maxwell, Timothy and Telleen-Lawton,
                   Timothy and Hume, Tristan and Hatfield-Dodds, Zac and Kaplan,
                   Jared and Brauner, Jan and Bowman, Samuel R and Perez, Ethan",
  month         =  "16~" # jul,
  year          =  2023,
  url           = "http://arxiv.org/abs/2307.13702",
  archivePrefix = "arXiv",
  primaryClass  = "cs.AI",
  eprint        = "2307.13702"
}

@MISC{google-2025-gemini,
  title        = "{Gemini 2.5: Our most intelligent AI model}",
  author       = "{Google}",
  month        =  "25~" # mar,
  year         =  2025,
  howpublished = "\url{https://blog.google/technology/google-deepmind/gemini-model-thinking-updates-march-2025/}",
  note         = "Accessed: 2025-11-10",
  language     = "en"
}

@MISC{zhou-2023-instruction-following,
  title         = "{Instruction-following evaluation for Large Language Models}",
  author        = "Zhou, Jeffrey and Lu, Tianjian and Mishra, Swaroop and
                   Brahma, Siddhartha and Basu, Sujoy and Luan, Yi and Zhou,
                   Denny and Hou, Le",
  month         =  "14~" # nov,
  year          =  2023,
  url           = "http://arxiv.org/abs/2311.07911",
  archivePrefix = "arXiv",
  primaryClass  = "cs.CL",
  eprint        = "2311.07911",
  doi           = "10.48550/arXiv.2311.07911"
}

@INPROCEEDINGS{tanzer-2023-benchmark-a,
  title     = "{A Benchmark for Learning to Translate a New Language from One
               Grammar Book}",
  author    = "Tanzer, Garrett and Suzgun, Mirac and Visser, Eline and Jurafsky,
               Dan and Melas-Kyriazi, Luke",
  booktitle = "{The Twelfth International Conference on Learning
               Representations}",
  month     =  "13~" # oct,
  year      =  2023,
  url       = "https://openreview.net/forum?id=tbVWug9f2h"
}

@INPROCEEDINGS{merrill-2024-expressive,
  title     = "{The Expressive Power of Transformers with Chain of Thought}",
  author    = "Merrill, William and Sabharwal, Ashish",
  booktitle = "{The Twelfth International Conference on Learning
               Representations}",
  year      =  2024,
  url       = "https://openreview.net/forum?id=NjNGlPh8Wh"
}

@INPROCEEDINGS{wang-2024-mmlu-pro,
  title     = "{MMLU-Pro: A More Robust and Challenging Multi-Task Language
               Understanding Benchmark}",
  author    = "Wang, Yubo and Ma, Xueguang and Zhang, Ge and Ni, Yuansheng and
               Chandra, Abhranil and Guo, Shiguang and Ren, Weiming and Arulraj,
               Aaran and He, Xuan and Jiang, Ziyan and Li, Tianle and Ku, Max
               and Wang, Kai and Zhuang, Alex and Fan, Rongqi and Yue, Xiang and
               Chen, Wenhu",
  booktitle = "{The Thirty-eight Conference on Neural Information Processing
               Systems Datasets and Benchmarks Track}",
  month     =  "13~" # nov,
  year      =  2024,
  url       = "https://openreview.net/forum?id=y10DM6R2r3#discussion"
}

@ARTICLE{chomsky-1959-certain,
  title     = "{On certain formal properties of grammars}",
  author    = "Chomsky, Noam",
  journal   = "Information and control",
  publisher = "Elsevier BV",
  volume    =  2,
  number    =  2,
  pages     = "137--167",
  month     =  "1~" # jun,
  year      =  1959,
  url       = "http://dx.doi.org/10.1016/S0019-9958(59)90362-6",
  doi       = "10.1016/s0019-9958(59)90362-6",
  issn      = "0019-9958,1878-2981",
  language  = "en"
}

@INPROCEEDINGS{pfau-2024-letS,
  title     = "{Let’s Think Dot by Dot: Hidden computation in transformer
               language models}",
  author    = "Pfau, Jacob and Merrill, William and Bowman, Samuel R",
  booktitle = "{First Conference on Language Modeling}",
  month     =  "26~" # aug,
  year      =  2024,
  url       = "https://openreview.net/forum?id=NikbrdtYvG"
}

@INPROCEEDINGS{shojaee-2025-illusion-a,
  title     = "{The Illusion of Thinking: Understanding the Strengths and
               Limitations of Reasoning Models via the Lens of Problem
               Complexity}",
  author    = "Shojaee, Parshin and Mirzadeh, Seyed Iman and Alizadeh, Keivan
               and Horton, Maxwell and Bengio, Samy and Farajtabar, Mehrdad",
  booktitle = "{The Thirty-ninth Annual Conference on Neural Information
               Processing Systems}",
  month     =  "29~" # oct,
  year      =  2025,
  url       = "https://openreview.net/forum?id=YghiOusmvw&referrer=%5Bthe%20profile%20of%20Samy%20Bengio%5D(%2Fprofile%3Fid%3D~Samy_Bengio1)"
}

@TECHREPORT{openai-2025-gpt-5,
  title       = "{GPT-5 System Card}",
  author      = "{OpenAI}",
  institution = "OpenAI",
  month       =  "13~" # aug,
  year        =  2025,
  url         = "https://cdn.openai.com/gpt-5-system-card.pdf",
  language    = "en"
}

@MISC{anthropic-2025-claudes,
  title        = "{Claude's extended thinking}",
  author       = "{Anthropic}",
  month        =  "24~" # feb,
  year         =  2025,
  howpublished = "\url{https://www.anthropic.com/news/visible-extended-thinking}",
  note         = "Accessed: 2025-12-5",
  language     = "en"
}

@INPROCEEDINGS{lambert-2025-tulu,
  title     = "{Tulu 3: Pushing Frontiers in Open Language Model Post-Training}",
  author    = "Lambert, Nathan and Morrison, Jacob and Pyatkin, Valentina and
               Huang, Shengyi and Ivison, Hamish and Brahman, Faeze and Miranda,
               Lester James Validad and Liu, Alisa and Dziri, Nouha and Lyu,
               Xinxi and Gu, Yuling and Malik, Saumya and Graf, Victoria and
               Hwang, Jena D and Yang, Jiangjiang and Le Bras, Ronan and
               Tafjord, Oyvind and Wilhelm, Christopher and Soldaini, Luca and
               Smith, Noah A and Wang, Yizhong and Dasigi, Pradeep and
               Hajishirzi, Hannaneh",
  booktitle = "{Second Conference on Language Modeling}",
  month     =  "26~" # aug,
  year      =  2025,
  url       = "https://openreview.net/forum?id=i1uGbfHHpH"
}

@MISC{yang-2025-ifevalcode,
  title         = "{IFEvalCode: Controlled Code Generation}",
  author        = "Yang, Jian and Zhang, Wei and Liu, Shukai and Chai, Linzheng
                   and Tan, Yingshui and Liu, Jiaheng and Zhang, Ge and Zhou,
                   Wangchunshu and Niu, Guanglin and Li, Zhoujun and Hui,
                   Binyuan and Lin, Junyang",
  month         =  "1~" # aug,
  year          =  2025,
  url           = "http://arxiv.org/abs/2507.22462",
  archivePrefix = "arXiv",
  primaryClass  = "cs.CL",
  eprint        = "2507.22462",
  doi           = "10.48550/arXiv.2507.22462"
}

@INPROCEEDINGS{yang-2023-large,
  title     = "{Large Language Models as Optimizers}",
  author    = "Yang, Chengrun and Wang, Xuezhi and Lu, Yifeng and Liu, Hanxiao
               and Le, Quoc V and Zhou, Denny and Chen, Xinyun",
  booktitle = "{The Twelfth International Conference on Learning
               Representations}",
  month     =  "13~" # oct,
  year      =  2023,
  url       = "https://openreview.net/forum?id=Bb4VGOWELI"
}

@INPROCEEDINGS{butoi-2025-training,
  title     = "{Training Neural Networks as Recognizers of Formal Languages}",
  author    = "Butoi, Alexandra and Khalighinejad, Ghazal and Svete, Anej and
               Valvoda, Josef and Cotterell, Ryan and DuSell, Brian",
  booktitle = "{The Thirteenth International Conference on Learning
               Representations}",
  year      =  2025,
  url       = "https://openreview.net/forum?id=aWLQTbfFgV"
}

@INPROCEEDINGS{turpin-2023-language-a,
  title     = "{Language Models Don't Always Say What They Think: Unfaithful
               Explanations in Chain-of-Thought Prompting}",
  author    = "Turpin, Miles and Michael, Julian and Perez, Ethan and Bowman,
               Samuel R",
  booktitle = "{Thirty-seventh Conference on Neural Information Processing
               Systems}",
  year      =  2023,
  url       = "https://openreview.net/forum?id=bzs4uPLXvi"
}

@INPROCEEDINGS{zhang-2024-inftyBench,
  title     = "{$\infty${Bench}: Extending long context evaluation beyond 100K
               tokens}",
  author    = "Zhang, Xinrong and Chen, Yingfa and Hu, Shengding and Xu, Zihang
               and Chen, Junhao and Hao, Moo and Han, Xu and Thai, Zhen and
               Wang, Shuo and Liu, Zhiyuan and Sun, Maosong",
  booktitle = "{Proceedings of the 62nd Annual Meeting of the Association for
               Computational Linguistics (Volume 1: Long Papers)}",
  publisher = "Association for Computational Linguistics",
  address   = "Stroudsburg, PA, USA",
  pages     = "15262--15277",
  year      =  2024,
  url       = "https://aclanthology.org/2024.acl-long.814/",
  doi       = "10.18653/v1/2024.acl-long.814"
}

@INPROCEEDINGS{finlayson-2022-what-a,
  title     = "{What makes instruction learning hard? An investigation and a new
               challenge in a synthetic environment}",
  author    = "Finlayson, Matthew and Richardson, Kyle and Sabharwal, Ashish and
               Clark, Peter",
  booktitle = "{Proceedings of the 2022 Conference on Empirical Methods in
               Natural Language Processing}",
  publisher = "Association for Computational Linguistics",
  address   = "Stroudsburg, PA, USA",
  pages     = "414--426",
  month     =  dec,
  year      =  2022,
  url       = "http://dx.doi.org/10.18653/v1/2022.emnlp-main.27",
  doi       = "10.18653/v1/2022.emnlp-main.27"
}

@INPROCEEDINGS{hsieh-2024-ruler,
  title     = "{RULER: What's the Real Context Size of Your Long-Context
               Language Models?}",
  author    = "Hsieh, Cheng-Ping and Sun, Simeng and Kriman, Samuel and Acharya,
               Shantanu and Rekesh, Dima and Jia, Fei and Ginsburg, Boris",
  booktitle = "{First Conference on Language Modeling}",
  year      =  2024,
  url       = "https://openreview.net/forum?id=kIoBbc76Sy"
}

@INPROCEEDINGS{arora-2024-zoology,
  title     = "{Zoology: Measuring and Improving Recall in Efficient Language
               Models}",
  author    = "Arora, Simran and Eyuboglu, Sabri and Timalsina, Aman and
               Johnson, Isys and Poli, Michael and Zou, James and Rudra, Atri
               and Re, Christopher",
  booktitle = "{The Twelfth International Conference on Learning
               Representations}",
  year      =  2024,
  url       = "https://openreview.net/forum?id=LY3ukUANko"
}

\clearpage
\appendix

\section{Grammar and String Novelty} \label{sec:grammar-growth}

The grammars and candidate strings generated by RELIC are almost certainly novel, meaning that they are very unlikely to have been generated and observed before during training.

\begin{lemma}
Let $G$ be a context-free grammar in Chomsky normal form with $n$ nonterminal symbols and $\abs{\Sigma} = t$ terminal symbols, and let $\ell$ be the length of a string drawn from $\Sigma^+$. Then there are at least $2^{n^3 + nt - 2n}$ many distinct grammars and $t^\ell$ many distinct strings.
\end{lemma}

\begin{proof}
To establish a lower bound on the number of reduced CFGs in Chomsky normal form we exhibit a set of such grammars and count them. Fix $V = \{\tok{NT}_1, \dotsc, \tok{NT}_n\}$ and $\Sigma = \{\tok{t}_1, \dotsc, \tok{t}_t\}$. Let $G'$ be a base grammar whose nonlexical productions are all rules of the form
\[
\{S \to \tok{NT}_i ~~ \tok{NT}_1\} \quad \text{for $1 \leq i \leq n$}, \label{eq:access}\tag{$\alpha$}
\]
and whose lexical productions are all rules of the form
\[
\{\tok{NT}_i \to \texttt{`$\tok{t}_1$'}\} \quad \text{for $1 \leq i \leq n$}. \label{eq:prod}\tag{$\beta$}
\]
It is clear that $G'$ is reduced, since every nonterminal $\tok{NT}_i$ is accessible from the $i$th nonlexical rule by \eqref{eq:access} and productive in the $i$th lexical rule by \eqref{eq:prod}. Furthermore, $G'$ remains reduced after adding any other lexical or nonlexical production rules since their inclusion does not affect productivity or accessibility. A lower bound on the number of reduced CNF CFGs is then the number of distinct grammars which may extend $G'$. There are $n^3$ possible distinct nonterminal production rules and $nt$ possible distinct terminal production rules. Since $G'$ contains $2n$ production rules, there are $n^3 + nt - 2n$ remaining distinct production rules to choose from, and so there are at least $2^{n^3 + nt - 2n}$ distinct reduced CNF CFGs.

The number of distinct strings on $t$ symbols of length $\ell$ is trivial. \qedhere
\end{proof}

\begin{remark}
The number of distinct grammar--sample pairs for fixed $n$, $t$, and $\ell$ is multiplicative:
\[
N_\text{pairs} \geq 2^{n^3+nt-2n} \cdot t^\ell.
\]
This grows \emph{very} fast. In practice, the space of distinct grammars and samples is extraordinarily large, and the chance of accidentally generating a grammar--sample pair which had been observed during training is vanishingly small.
\end{remark}

\section{Dataset Information} \label{sec:grammar-stats}

\subsection{Distributional Statistics for Grammars and Strings}

To construct the static evaluation set used in our experiments, we first over-generate grammars ($\sim$1000 total) where the four generating hyperparameters of each grammar (the numbers of terminal $\Nterm$ and nonterminal $\Nnonterm$ symbols and the numbers of lexical $\Nlex$ and nonlexical $\Nnonlex$ production rules) are bounded above by 500. We then subsample these grammars to include the 200 grammars whose hyperparameters are minimally correlated with one another. \Cref{fig:grammar-correlations} below reports the hyperparameter correlations in the released set. Note that the correlations between $\Nterm\sim\Nlex$ and $\Nnonterm\sim\Nnonlex$ are higher than the others since the former term is bounded above by the latter in both cases.

\begin{table}[h]
  \centering
  \setlength{\tabcolsep}{10pt}
  \small
  \begin{tabularx}{\linewidth}{XSSS}
    \toprule
    & $n_{\text{term}}$ & $n_{\text{nonterm}}$ & $n_{\text{lex}}$  \\
    \midrule
    $n_{\text{nonterm}}$
      &  0.01 & & \\
    $n_{\text{lex}}$
      &  0.54 &  0.07 & \\
    $n_{\text{nonlex}}$
      &  -0.00 &  0.31 &  -0.01 \\
      \bottomrule
  \end{tabularx}
      \caption{Correlations between generating hyperparameters for the released static set. Note that $\Nlex$ and $\Nterm$ are inherently correlated.}
    \label{fig:grammar-correlations}
\end{table}

From each grammar we sample positive and negative strings. Positive strings are sampled by converting each grammar into a probabilistic context-free grammar \citep{booth-1973-applying} with uniform probability for each right-hand-side path among all productions which share a left-hand-side nonterminal and sampling a production stochastically. We over-sample positive strings and filter so that they are of length at most $50$, and such that we have no more than $10$ strings per length. Negative strings are sampled by drawing strings over the set of terminal symbols $\Sigma^+$ of fixed length $1 \leq \ell \leq 50$ uniformly-at-random and rejecting any strings which are parseable by the grammar. We repeat this process until we have $10$ strings per length. 

\begin{figure}
    \centering
    \includegraphics[width=\linewidth]{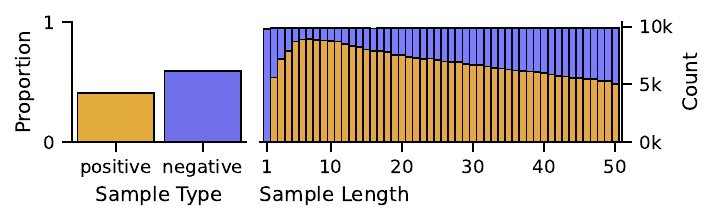}
    \caption{Proportions of example types represented in the static dataset, in aggregate (\textbf{left}) and broken down by example length (\textbf{right}).}
    \label{fig:sample-stats}
\end{figure}

Since positive examples are not drawn with a pre-determined length and not all grammars can generate $10$ strings for each length, the resulting set of sampled strings will in some cases be smaller than that of the negative examples; \cref{fig:sample-stats} shows the relative proportions of positive and negative samples drawn from the released set of grammars. For our evaluations, we choose not to post-hoc rebalance example types for each length since the distribution of positive examples by length is a property of the grammar. Since not all grammars will generate strings of every length in equal proportions, the length of an example contains relevant information about the example's type, albeit information which is not provided to the model independently from the grammar itself. For a model to justifiably use the example length to arrive at the correct answer, it must derive the relevant properties from the production rules itself.

\begin{figure}
    \centering
    \includegraphics[width=\linewidth]{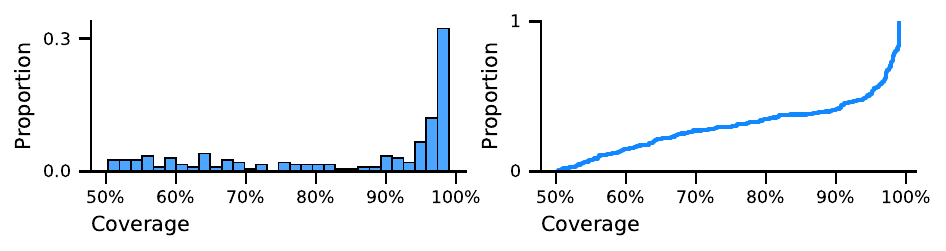}
    \caption{Distribution of grammars by coverage (i.e., the size of the language they generate, measured as the number of positive examples of lengths $\ell \leq 50$, with a maximum of \num{10} examples/length, out of a theoretical maximum of \num{500}) shown as a histogram (\textbf{top}) and a cumulative distribution function (\textbf{bottom}).}
    \label{fig:coverage}
\end{figure}

We refer to the number of examples for a grammar, relative to the theoretical maximum of $1000$ strings (= 2 example types * 50 lengths * 10 examples per length per type) as the grammar's \emph{coverage}; \cref{fig:coverage} shows the coverage of the grammars used in our experiments.

\subsection{Why are \texorpdfstring{$\Nterm$}{Nterm} and \texorpdfstring{$\Nlex$}{Nlex} Correlated?}
\label{sec:term_lex_correlations}

In any context-free grammar the number of terminal symbols $\Nterm$ is bounded above by the number of lexical production rules $\Nlex$; to see why, consider the following minimal grammar with a single lexical rule and a single terminal symbol:
\[
\begin{matrix*}[l]
    \tok{S} \to \tok{A A} \\
    \tok{A} \to \verb|'a'|
\end{matrix*}
\]
To add a new terminal symbol \verb|'b'| to the grammar, we must add a production rule which yields it:
\[
\begin{matrix*}[l]
    \tok{S} \to \tok{A A} \\
    \tok{A} \to \verb|'a'| \\
    \tok{A} \to \verb|'b'|
\end{matrix*}
\]
Thus, $\Nterm \leq \Nlex$. To see how it could be lower, consider a grammar like the following where multiple distinct nonterminals yield the same terminal:
\[
\begin{matrix*}[l]
    \tok{S} \to \tok{A B} \\
    \tok{A} \to \verb|'a'| \\
    \tok{B} \to \verb|'a'|
\end{matrix*}
\]
Here the grammar has two lexical production rules but only a single terminal symbol. A similar argument extends to the relationship between $\Nnonterm$ and $\Nnonlex$.

\section{Regressions \& Correlation Data} 
\label{sec:results-tables}

\Cref{tab:corr} reports the Pearson correlation coefficients for model accuracy and macro F1 score with the log transforms of the four generating hyperparameters of each grammar (the numbers of terminal $\Nterm$ and nonterminal $\Nnonterm$ symbols and the numbers of lexical $\Nlex$ and nonlexical $\Nnonlex$ production rules).

\begin{table*}[ht!]
    \footnotesize
    \centering
    \begin{tabularx}{\textwidth}{XcSSSS} \toprule
         & & {$\log(\Nterm)$} & {$\log(\Nlex)$} & {$\log(\Nnonterm)$} & {$\log(\Nnonlex)$} \\
         \midrule
        \multirow{ 2}{*}{\texttt{gpt-4.1-nano}} & $r_\text{F1}$ & 0.26 & 0.23 & 0.14 & -0.10 \\ 
        & $r_\text{Acc.}$ & 0.15 & 0.17 & 0.09 & -0.05 \\ \midrule
        \multirow{ 2}{*}{\texttt{gpt-4.1-mini}} & $r_\text{F1}$ & 0.07 & -0.02 & -0.13 & -0.57 \\
        & $r_\text{Acc.}$ & 0.09 & 0.02 & -0.13 & -0.45 \\ \midrule
        \multirow{ 2}{*}{\texttt{gpt-4.1}} & $r_\text{F1}$ & -0.11 & -0.37 & -0.52 & -0.59 \\
        & $r_\text{Acc.}$ & -0.13 & -0.28 & -0.25 & -0.27 \\ \midrule
        \multirow{ 2}{*}{\texttt{o4-mini}} & $r_\text{F1}$ & -0.20 & -0.39 & -0.60 & -0.79 \\
        & $r_\text{Acc.}$ & -0.15 & -0.28 & -0.20 & -0.36 \\ \midrule
        \multirow{ 2}{*}{\texttt{o3}} & $r_\text{F1}$ & -0.12 & -0.22 & -0.37 & -0.75 \\
        & $r_\text{Acc.}$ & -0.06 & -0.12 & -0.17 & -0.40 \\ \midrule
        \multirow{ 2}{*}{\texttt{gemma-3-1b}} & $r_\text{F1}$ & -0.21 & -0.24 & -0.19 & -0.20 \\
        & $r_\text{Acc.}$ & -0.09 & -0.13 & -0.13 & -0.11 \\ \midrule
        \multirow{ 2}{*}{\texttt{gemma-3-4b}} & $r_\text{F1}$ & -0.22 & -0.36 & -0.49 & -0.61 \\
        & $r_\text{Acc.}$ & -0.05 & -0.09 & -0.21 & -0.19 \\ \midrule
        \multirow{ 2}{*}{\texttt{DSR1-7B}} & $r_\text{F1}$ & -0.14 & -0.24 & -0.43 & -0.38 \\
        & $r_\text{Acc.}$ & -0.12 & -0.20 & -0.20 & -0.20
        \\ \bottomrule
    \end{tabularx}
    \caption{Pearson correlation coefficients $r$ between models' accuracy/macro F1 scores and grammar hyperparameters, including the numbers of terminal symbols $\Nterm$, nonterminal symbols $\Nnonterm$, lexical productions $\Nlex$, and nonlexical productions $\Nnonlex$. Correlation scores are taken over the mean F1 and accuracy values grouped by grammar.}
    \label{tab:corr}
\end{table*}

We also report the coefficients of regression for accuracy (as a percentage) versus model type, $\Nnonlex$, and example length $\ell$ in~\cref{tab:regression}. Both $\Nnonlex$ and $\ell$ were log-transformed and centered before being entered into the regression.

\begin{table}[ht!]
    \centering \footnotesize
    \begin{tabularx}{\linewidth}{
        X
        S[round-mode=places, round-precision=2]
        @{}S[round-mode=places, round-precision=3]
        @{\hskip 0.6em}l
    }
         \toprule
         \textbf{Term} & \textbf{Coeff.} & \textbf{$p$-value} & \textbf{Sig.} \\ \midrule
         Intercept & 46.5642 & {$<$ 0.001} & \texttt{***} \\
         \quad \texttt{gpt-4.1-mini} & 13.1801 & {$<$ 0.001} & \texttt{***} \\
         \quad \texttt{gpt-4.1} & 19.4300 & {$<$ 0.001} & \texttt{***} \\
         \quad  \texttt{o4-mini} & 21.6878 & {$<$ 0.001} & \texttt{***} \\
         \quad \texttt{o3} & 26.2557 & {$<$ 0.001} & \texttt{***} \\
         \quad \texttt{gemma-3-1b} & 7.8714 & {$<$ 0.001} & \texttt{***} \\
         \quad \texttt{gemma-3-4b} & 7.3847 & {$<$ 0.001} & \texttt{***} \\
         \quad \texttt{DSR1-7B} & 9.6567 & {$<$ 0.001} & \texttt{***} \\ \midrule
         $\log_{10}(\Nnonlex) * \texttt{gpt-4.1-nano}$ & -2.2806 & {$<$ 0.001} & \texttt{***} \\
         $\log_{10}(\Nnonlex) * \texttt{gpt-4.1-mini}$ & -24.1731 & {$<$ 0.001} & \texttt{***} \\
         $\log_{10}(\Nnonlex) * \texttt{gpt-4.1}$ & -10.2849 & {$<$ 0.001} & \texttt{***} \\
         $\log_{10}(\Nnonlex) * \texttt{o4-mini}$ & -14.4147 & {$<$ 0.001} & \texttt{***} \\
         $\log_{10}(\Nnonlex) * \texttt{o3}$ &-23.5386 & {$<$ 0.001} & \texttt{***} \\
         $\log_{10}(\Nnonlex) * \texttt{gemma-3-1b}$ & -4.4397 & {$<$ 0.001} & \texttt{***} \\
         $\log_{10}(\Nnonlex)* \texttt{gemma-3-4b}$ &-9.9279 & {$<$ 0.001} & \texttt{***} \\
         $\log_{10}(\Nnonlex) * \texttt{DSR1-7B}$ & -9.7763 & {$<$ 0.001} & \texttt{***} \\ \midrule
         $\log_{10}(\ell) * \texttt{gpt-4.1-nano}$ & -26.7932 & {$<$ 0.001} & \texttt{***} \\
         $\log_{10}(\ell) * \texttt{gpt-4.1-mini}$ & 0.7205 & 0.385 & \texttt{} \\
         $\log_{10}(\ell) * \texttt{gpt-4.1}$ & 15.9320 & {$<$ 0.001} & \texttt{***} \\
         $\log_{10}(\ell) * \texttt{o4-mini}$ & 8.7292 & {$<$ 0.001} & \texttt{***} \\
         $\log_{10}(\ell) * \texttt{o3}$ & -1.3561 & 0.102 & \\
         $\log_{10}(\ell) * \texttt{gemma-3-1b}$ & 28.3663 & {$<$ 0.001} & \texttt{***} \\
         $\log_{10}(\ell) * \texttt{gemma-3-4b}$ & 41.1999 & {$<$ 0.001} & \texttt{***} \\
         $\log_{10}(\ell) * \texttt{DSR1-7B}$ & 28.0295 & {$<$ 0.001} & \texttt{***} \\ \midrule
         $\log_{10}(\Nnonlex) * \log_{10}(\ell)$ & -10.2807 & {$<$ 0.001} & \texttt{***}
         \\ \bottomrule
    \end{tabularx}
    \caption{Regression coefficients for accuracy by model, $\log_{10}(\Nnonlex)$, and example length $\ell$. The base level of the model factor is \texttt{gpt-4.1-nano}. Significance values are listed as \texttt{***} for $p < 0.001$, \texttt{**} for $p < 0.01$, and \texttt{*} for $p < 0.05$.}
    \label{tab:regression}
\end{table}

\begin{figure*}
    \centering
    \includegraphics[width=\linewidth]{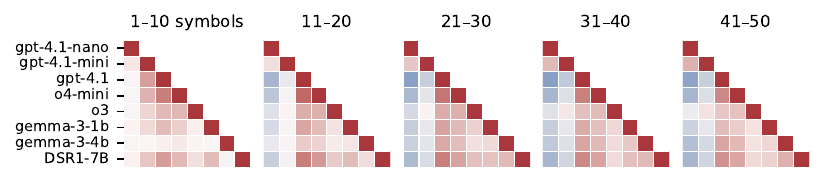}
    \includegraphics[width=\linewidth]{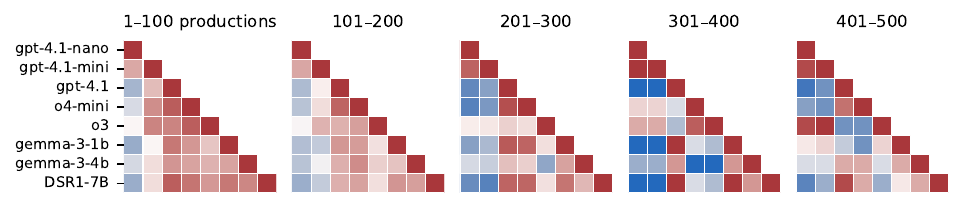}
    \caption{Spearman's rank correlation coefficients for accuracies between different models (\textbf{top:} per-example, faceted by example length; \textbf{bottom:} per-grammar, faceted by number of non-lexical productions).}
    \label{fig:sample-rank}
\end{figure*}

\Cref{fig:sample-rank} shows the full Spearman's rank correlation coefficients across models for the difficulties of individual grammars and samples.

\section{Inference Cost, Setup, and Hyperparameters} \label{sec:hyps}

Evaluations on OpenAI models used roughly \$15k total of compute credits. Evaluations on the open-weight models were run on a GPU cluster using NVIDIA A100s, H100s, V100s, and RTX8000s; models were loaded using the Hugging Face \texttt{transformers} library \citep{wolf-2020-transformers}. 

\Cref{tab:hyps} reports the inference hyperparameters used in our experiments. For the open-weight models, which we run on local hardware, we restrict the number of new tokens (\verb|max_completion_tokens|) to \si{4096} to limit memory usage and inference time. For all models, we generate completions with sampling using the default parameters (temperature $\tau$ and nucleus constant $p$) specified by the model or API.

\begin{table}[ht!]
    \centering \footnotesize
    \begin{tabularx}{\linewidth}{XcS@{}S}
        \toprule
        \textbf{Model} & \textbf{Token limit} & $\tau$ & $p$ \\
        \midrule
         \texttt{gpt-4.1-nano} & {None} & 1.0 & 1.00  \\
         \texttt{gpt-4.1-mini} & {None} & 1.0 & 1.00  \\
         \texttt{gpt-4.1} & {None} & 1.0 & 1.00  \\ \midrule
         \texttt{o4-mini} & {None} & 1.0 & 1.00  \\ 
         \texttt{o3} & {None} & 1.0 & 1.00  \\ \midrule
         \texttt{gemma-3-1b-it} & 4096 & 1.0 & 0.95  \\
         \texttt{gemma-3-4b-it} & 4096 & 1.0 & 0.95  \\ \midrule
         \texttt{DeepSeek-R1-Distill-Qwen-7B} & 2048 & 0.6 & 0.95
         \\ \bottomrule
    \end{tabularx}
    \caption{Inference hyperparameters used in experiments.}
    \label{tab:hyps}
\end{table}

\section{Chains-of-Thought}

This appendix provides a few examples of the chains-of-thought (tokens generated between the prompt and the answer) produced by \texttt{DeepSeek-R1-Distill-Qwen-7B} (which we abbreviate here as \texttt{DSR1}).

\subsection{Example A: Faulty Reasoning} \label{sec:dsr1-20250218222557-a}

In the example shown below, \texttt{DSR1} reports identifying a strategy that, while vaguely described, is consistent with the appropriate way to determine if the example is generated by the grammar (\textit{``start from the string and see if I can break it down into the nonterminals according to the production rules''}; see \parref{ex:dsr1-parse-strat}). It begins exploring this strategy, but quickly backtracks and veers off into a heuristic and irrelevant analysis based on string length (\textit{``Wait, let me count the tokens''}; \parref{ex:dsr1-counting}). It then backtracks again, and attempts to find a derivation for the first symbol in the example~(\parref{ex:dsr1-first-symbol}). It correctly identifies two production rules that together can generate the first symbol in the example (the first symbol is \texttt{t75}, and the relevant rules are \texttt{\{S $\to$ NT97 NT1, NT97 $\to$ `t75'\}}). It then proceeds to determine how the second symbol in the example, \texttt{t104}, could be derived. It notices that \texttt{NT1}---the second nonterminal in the rule \texttt{S $\to$ NT97 NT1}, which it used to derive the first symbol---does not directly produce \texttt{t104}, and considers this a potential reason for rejecting the string~(\parref{ex:dsr1-second-symbol}). It then considers the possibility that \texttt{NT1} produces \texttt{t104} not directly but through intermediate productions~(\parref{ex:dsr1-nonimmediate}); it then rejects this possibility based on the presence of a different lexical production rule involving \texttt{NT1}---although this is an incorrect reason to reject this possibility---and concludes, incorrectly, that the example is not generated by the grammar~(\parref{ex:dsr1-wrong}).

\begin{llm}[linenumbered]
\faint{...} I think the approach is to start from the string and see if I can break it down into the nonterminals according to the production rules. \faint{...} \label{ex:dsr1-parse-strat}

\faint{... since the string has 41 symbols, which is an odd number.} Wait, let me count the tokens: \faint{the string is "t75 t104 t120 t27 t137 t27 t11 t11 t11 t173 t181 t27 t84 t11 t82 t104 t120 t168 t69 t141 t158 t9 t12 t140 t102 t114 t190 t141 t147 t79 t141 t159 t178 t183 t183 t83 t141 t158 t9 t55 t74 t170 t140 t102 t114 t190".} Let me count these: 41 tokens. \faint{...} \label{ex:dsr1-counting}

Alternatively, starting from the beginning, the first symbol is `t75', which must come from some terminal in the grammar. Looking at the grammar, let's find which nonterminals have a production ending with `t75'. \faint{Wait, in the grammar, NT44 -> `t132', and NT197 -> `t178', and NT97 -> `t75' according to the end of the grammar list: NT97 -> `t75'.} So, NT97 can directly produce `t75'. \label{ex:dsr1-first-symbol}

Now, looking at NT1: according to the grammar, NT1 -> `t1'. So, that's a terminal. So the first two symbols in the string must be `t75' followed by `t1'. Let me check the given string: the first two tokens are `t75 t104', but wait, that's not `t75' followed by `t1'. So that's a problem. \label{ex:dsr1-second-symbol}

Wait, alternatively, maybe the NT1 is not the immediate right child but another expansion. \faint{...} \label{ex:dsr1-nonimmediate}

Therefore, it's not possible that the entire string is generated by this grammar because the second symbol doesn't match `t1'. \faint{...} \label{ex:dsr1-wrong}
\end{llm}

\vspace*{0em}

\subsection{Example B: Irrelevant Distractions} \label{sec:dsr1-20250402155408_676876-a}

Here, \texttt{DSR1} is prompted by a (negative) example consisting of \textit{a single symbol} \texttt{t136}; though it identifies that this token is derivable by some lexical production rule, it immediately begins investigating irrelevant derivations which contain additional terminal symbols~(\parref{ex:dsr1-irrelevant-info}). It continues along this line before arguing (incorrectly) that the example seems plausible since it seems like there exist derivations containing the provided terminal symbol~(\parref{ex:dsr1-wrong-again}).

\begin{llm}[linenumbered]
\llmfiller

\faint{Looking at the patterns, I see that `t136' might correspond to multiple parts.} I remember that `t136' could link to `t39' and `t6' based on some earlier patterns. \boxpara\label{ex:dsr1-irrelevant-info}
\llmfiller
\faint{Putting it all together,} it seems like `t136' is a valid node in the context. \faint{So, the answer should be `Yes'.} \boxpara\label{ex:dsr1-wrong-again}
\end{llm}

\vspace*{0em}

\subsection{Example C: Incorrect Strategy} \label{sec:dsr1-20250402155408_676876-b}

Here, \texttt{DSR1} adopts a strategy that is entirely irrelevant: it decides to check if the string contains the symbol \texttt{t27}~(\parref{ex:dsr1-t27}), a symbol which not only does not appear in the example it was given, but does not even appear in the grammar in question. It then proceeds to check if any symbol in the provided string is \texttt{t27} before rejecting the example because \texttt{t27} was not present~(\parref{ex:dsr1-wrong-27}). Because the example in this case happened to be a negative example, \texttt{DSR1} happened to produce the correct answer, but it reached this answer for irrelevant reasons.

\begin{llm}[linenumbered]
\llmfiller
First, I see that the task involves checking if the token `t27' appears in the output of a certain string evaluation. \faint{The string to evaluate is `t44 t249 t80 t333 t171 t194 t329 t224 t196 t64 t8 t43 t16 t15 t280 t220 t107 t230 t334 t207 t19 t296'.} \boxpara\label{ex:dsr1-t27}
\llmfiller
So, based on my evaluation, the output of the evaluation doesn't contain `t27'. Therefore, the condition isn't met, and the answer should be `No'. \boxpara\label{ex:dsr1-wrong-27}
\end{llm}

\vspace*{0em}

\section{Strategy Classification Details} \label{sec:llm-judge}

To categorize the strategies used by the \texttt{gpt-4.1-} series of models, we employ the `LLM-as-judge' framework \cite{zheng-2023-judging}: we prompt a judge model to classify other models' responses according to a rubric. To validate this methodology, we collect human judgments on model strategies for 62 different completions, balanced for binned sample length (short, medium, and long) and sample class (positive or negative). Human judgments are provided by paper authors. We then select two candidate judge LLMs, \texttt{o4-mini} and \texttt{gpt-5}, and prompt these judge models to categorize the same 62 samples. We compare the majority-category selected by the human annotators to that of the LLMs to compute the validation agreement between human and LLM judgment. We then select the judge LLM with the highest validation agreement to classify model strategies for a much larger subsampling of the full data, and report these values in~\cref{fig:gpt-strategy}. We prompt all judge models with the same guidelines given to human annotators, shown below:

\begin{prompt}
You will be presented with a completion from an LLM which was given a context-free grammar and a string of symbols drawn from that grammar's set of terminal symbols and asked to determine whether the string is generated by the grammar or not. Your job is to classify how the LLM attempted to solve the task by binning the completion strategy into one of the following categories:

- \verb|`heuristic`|: The LLM attempts to solve the task by using heuristics it surmises from the grammar, such as “if the string is long, it is likely generated by the grammar” or “the string only contains terminal symbols present in the grammar, so it’s likely a positive sample”. Count strategies as heuristic if they appeal to the existence of certain production rules but do not rigorously determine that no such derivation exists.\\
- \verb|`rule-based`|: The LLM attempts to solve the task by writing out the FULL DERIVATION of the sample from the grammar, or rigorously determining that no such derivation exists. Only count strategies as rule-based if the LLM doesn’t use any guesswork to reach its final conclusion.\\
- \verb|`code`|: The LLM attempts to solve the task by writing out a program or algorithm which it claims will solve the task. This includes writing out a program in a programming language, or writing out pseudocode.

You can write as much as you want in your answer, but please end your response with the name of the classification you think is most appropriate.

Here is the LLM's completion:

\begin{verbatim}
```
{completion}
```
\end{verbatim}
\end{prompt}

\vspace*{0em}

We found that \texttt{o4-mini} had a human-agreement of 68\% while \texttt{gpt-5} had a human-agreement of 83\%, and so report \texttt{gpt-5}'s analysis in the main text, though we note that \texttt{o4-mini}'s classifications generally agree with those of \texttt{gpt-5}. 

\section{Finetuning Details} \label{sec:ft}

We fine-tune \texttt{gemma-3-1b-it} and \texttt{gemma-3-4b-it} on a subset of RELIC (\cref{sec:ft-main}). We randomly select 100 grammars and generate an 80\%-20\% train-validation split, training on the RELIC prompt (see Prompt 1) in~\cref{sec:exp-setup} along with the correct response. We fine-tune each model for five epochs, after which point validation loss begins to diverge. For each model size we perform five training runs from different random seeds. We use the AdamW optimizer \citep{loshchilov-2019-decoupled} with standard parameters ($\beta_1 = 0.9$, $\beta_2 = 0.999$) and an initial learning rate of $5\times10^{-5}$, using the TRL framework from HuggingFace \citep{vonWerra-2020-trl}.

\end{document}